\documentclass[lettersize,journal]{IEEEtran}
\usepackage{amsfonts}

\usepackage{array}
\usepackage[caption=false,font=normalsize,labelfont=sf,textfont=sf]{subfig}
\usepackage{textcomp}
\usepackage{stfloats}
\usepackage{url}
\usepackage{verbatim}
\usepackage{graphicx}
\usepackage{cite}

\graphicspath{{figs/}}

\usepackage{makecell}
\usepackage{bm}

\usepackage{amsmath}
\usepackage{amssymb}

\DeclareMathOperator*{\argmax}{argmax}

\usepackage{algorithm}
\usepackage{algpseudocode}
\usepackage{bbding}

\usepackage{enumitem}

\usepackage{multicol}
\usepackage{multirow}
\usepackage{booktabs}

\usepackage{xcolor, array, colortbl}

\definecolor{deepred}{rgb}{0.698,0.133,0.133}
\definecolor{blue}{rgb}{0,0,1}

\definecolor{orange}{rgb}{1,0.38,0}
\definecolor{beige}{rgb}{0.639,0.58,0.502}

\definecolor{lightgray}{rgb}{.91,.91,.91}

\usepackage{tikz,xcolor,hyperref}
\definecolor{lime}{HTML}{A6CE39}
\DeclareRobustCommand{\orcidicon}{%
    \begin{tikzpicture}
    \draw[lime, fill=lime] (0,0) 
    circle [radius=0.16] 
    node[white] {{\fontfamily{qag}\selectfont \tiny ID}};    \draw[white, fill=white] (-0.0625,0.095) 
    circle [radius=0.007];    \end{tikzpicture}
    \hspace{-2mm}}
\foreach \x in {A, ..., Z}{%
    \expandafter\xdef\csname orcid\x\endcsname{\noexpand\href{https://orcid.org/\csname orcidauthor\x\endcsname}{\noexpand\orcidicon}}
    }

\begin{document}

\title{Exploring Stability-Plasticity Trade-offs for Continual Named Entity Recognition}

\author{Duzhen Zhang,
        Chenxing Li,
        Jiahua Dong, Qi Liu, and Dong Yu\orcidA{},~\IEEEmembership{Fellow,~IEEE}
        \thanks{Manuscript created November, 2024.}
       \thanks{Duzhen Zhang and Jiahua Dong are with the Mohamed bin Zayed University of Artificial Intelligence, Abu Dhabi, United Arab Emirates (E-mail: duzhen.zhang@mbzuai.ac.ae, dongjiahua1995@gmail.com).}
   
         \thanks{Chenxing Li is with the Tencent, AI Lab, Beijing, China (E-mail: chenxingli@tencent.com).}

    \thanks{Qi Liu is with the University of Hong Kong, Hong Kong, China (E-mail: liuqi@cs.hku.hk).}     
          \thanks{Dong Yu are with the Tencent, AI Lab, Bellevue, WA 98004 USA (E-mail: dyu@global.tencent.com).}
    \thanks{The corresponding authors are Chenxing Li and Dong Yu.}
    \thanks{This work was supported in part by the National Nature Science Foundation of China under Grant 62133005.}
        }

\markboth{Journal of \LaTeX\ Class Files,~Vol.~14, No.~8, August~2021}%
{Shell \MakeLowercase{\textit{et al.}}: A Sample Article Using IEEEtran.cls for IEEE Journals}

\maketitle

\begin{abstract}

Continual Named Entity Recognition (CNER) is an evolving field that focuses on sequentially updating an existing model to incorporate new entity types. Previous CNER methods primarily utilize Knowledge Distillation (KD) to preserve prior knowledge and overcome catastrophic forgetting, strictly ensuring that the representations of old and new models remain consistent. Consequently, they often impart the model with excessive stability (\emph{i.e.}, retention of old knowledge) but limited plasticity (\emph{i.e.}, acquisition of new knowledge). To address this issue, we propose a Stability-Plasticity Trade-off (SPT) method for CNER that balances these aspects from both representation and weight perspectives. From the representation perspective, we introduce a pooling operation into the original KD, permitting a level of plasticity by consolidating representation dimensions. From the weight perspective, we dynamically merge the weights of old and new models, strengthening old knowledge while maintaining new knowledge. During this fusion, we implement a weight-guided selective mechanism to prioritize significant weights.
Moreover, we develop a confidence-based pseudo-labeling approach for the current non-entity type, which predicts entity types using the old model to handle the semantic shift of the non-entity type, a challenge specific to CNER that has largely been ignored by previous methods.
Extensive experiments across ten CNER settings on three benchmark datasets demonstrate that our SPT method surpasses previous CNER approaches, highlighting its effectiveness in achieving a suitable stability-plasticity trade-off.

\end{abstract}


\section{Introduction}

\IEEEPARstart{N}{amed} Entity Recognition (NER) has become an essential area of research within information extraction \cite{zhang2024ssr,zhang2024cross}. The primary objective of NER is to classify each token in a sequence, determining whether it belongs to one of several predefined entity types or if it should be categorized as the non-entity type~\cite{ma2016end}. 
In recent years, the emergence of Pre-training Language Models (PLMs)~\cite{kenton2019bert} has revolutionized the field of NER, pushing it into a new era of capabilities and applications. Traditionally, NER has functioned under a paradigm where tokens are sorted into a set of fixed entity types—such as Organization, Person, and others—with the model being trained in a single, comprehensive learning process. 
However, real-world applications demand a more dynamic approach, where NER models must continuously identify and adapt to emerging entity types without requiring a full retraining cycle. This innovative approach is known as Continual Named Entity Recognition (CNER), and it has garnered substantial research interest due to its significant practical implications~\cite{monaikul2021continual,zheng2022distilling}. For example, voice-activated assistants like Siri and Alexa frequently need to recognize new entity types, such as Genre or Actor, to accurately interpret evolving user intents (\emph{e.g.}, GetMovie). This adaptability is critical for maintaining the relevance and accuracy of these systems in rapidly changing environments.

Similar to other challenges faced in the realm of continual learning, CNER also contends with the enduring issue of catastrophic forgetting~\cite{robins1995catastrophic,mccloskey1989catastrophic,goodfellow2013empirical,kirkpatrick2017overcoming,zhang2025federated}. Catastrophic forgetting refers to the tendency of neural networks to lose previously learned knowledge when they are exposed to and integrate new information. This issue is particularly pronounced in the context of CNER, where a model’s ability to identify previously learned entity types diminishes as it learns to recognize new ones. When the model updates its parameters to accommodate new entity types, the adjustments can inadvertently disrupt the existing knowledge base, leading to reduced accuracy or even complete forgetting of previously acquired information. This challenge becomes especially acute when models are trained sequentially on different entity types without implementing strategies specifically designed to mitigate forgetting. 

To counteract this, current CNER methods~\cite{monaikul2021continual,xia2022learn,zheng2022distilling,ma2023learning} frequently leverage Knowledge Distillation (KD)~\cite{hinton2015distilling}. KD involves extracting crucial representations, such as intermediate features and output probabilities, from the previous model and transferring them to the new model. This technique ensures a rigorous consistency between the representations of the old and new models, effectively preserving the learned knowledge from earlier stages. Consequently, while these methods significantly enhance the model's stability—by maintaining the integrity of old knowledge—they often come at the expense of the model’s plasticity, or its ability to acquire and integrate new knowledge. This trade-off between stability and plasticity presents a significant challenge in the development of more robust and flexible CNER systems, as it underscores the difficulty of simultaneously preserving old knowledge while efficiently learning and adapting to new information.

To tackle this issue, we introduce a Stability-Plasticity Trade-off (SPT) method for CNER. This approach aims to balance these aspects from both representation and weight perspectives. 
From a representation standpoint, we enhance the original KD method by incorporating a pooling operation. This modification strikes a careful balance between stability and plasticity by consolidating representation dimensions.
Regarding weights, we propose a strategy to fuse the weights of old and new models using a dynamic factor. This approach seeks to achieve a new equilibrium between past and current knowledge without adding significant computational overhead. To refine this integration process, we also introduce a weight-guided selective mechanism that discerns crucial weights. Initially, it evaluates the significance of model weights related to old entity types based on Fisher information and subsequently integrates these critical weights from both old and new models. 

Furthermore, we identify a unique issue specific to CNER: the semantic shift of the non-entity type. This suggests that the current non-entity type may encompass the genuine non-entity type, previously learned entity types, or future types not yet encountered. Without mechanisms to distinguish tokens related to past entity types from the true non-entity type, this semantic shift could exacerbate catastrophic forgetting. 
To address this challenge, we introduce a confidence-based pseudo-labeling strategy. This approach utilizes the old model to identify previous entity types within the current non-entity type and reduces recognition errors from the old model using entropy-based confidence thresholds.

In summary, the main contributions of this paper are:

\begin{itemize}
\item We introduce a SPT method for CNER that balances old and new knowledge from both representation and weight perspectives. In terms of representation, we incorporate pooling operations into the original KD, aiming to effectively mitigate catastrophic forgetting by consolidating representation dimensions properly.

\item As for weight, we dynamically merge the weights of old and new models to efficiently strike a new balance between previous and current entity types, thereby mitigating catastrophic forgetting. To avoid underfitting of new entity types, we implement a weight-guided selective mechanism. This mechanism selectively incorporates essential weights from the old model related to previous entity types into the new model.

\item Extensive experiments conducted across ten CNER settings using three benchmark datasets validate the effectiveness of our proposed SPT method. The results demonstrate that our approach outperforms several existing CNER methods and achieves new State-Of-The-Art (SOTA) performance in the field.

\end{itemize}

This study significantly extends our earlier EMNLP'2023 conference paper \cite{zhang2023continual}. The key enhancements of this paper, in contrast to \cite{zhang2023continual}, are outlined below:

\begin{itemize}
    \item Since improvements solely from the original representation perspective \cite{zhang2023continual} are limited, we explore stability-plasticity trade-offs from both representation and weight perspectives to enhance CNER performance. In this newly introduced weight perspective, we dynamically fuse the weights of old and new models to reinforce existing knowledge while integrating new information.

    \item To enhance this fusion process, we devise a weight-guided selective mechanism that integrates old and new model weights selectively, prioritizing weight importance concerning old entity types.

    \item Extensive experiments demonstrate that our SPT method further enhances the performance of our previously proposed CNER method \cite{zhang2023continual}. Moreover, we introduce additional quantitative experiments, incorporating finer-grained metrics to illustrate improvements in recognition of old and new entity types, as well as overall performance throughout successive CNER steps.
    
    \item We also analyze key hyper-parameters, investigate the stability of learning orders for entity types, and explore performance enhancements when using larger language models as encoder backbones. {Additionally, we conduct an efficiency analysis of the SPT method and further demonstrate its generalization capability on domain-specific Chinese datasets.}
\end{itemize}

\section{Related Work}
\label{Related_Work}

\subsection{Continual Learning}
\label{cl}

Continual learning encompasses the ability to acquire and retain knowledge across a sequence of tasks without compromising performance on previously learned tasks~\cite{chen2018lifelong,dong2022federated,dong2023federated,wang2024comprehensive,zheng2024towards,dong2024continually,zheng2025lifelong}. To address the challenges inherent in continual learning, existing methods can be broadly classified into three categories: memory-based, dynamic architecture-based, and regularization-based approaches. 
Memory-based methods~\cite{lopez2017gradient,rebuffi2017icarl,shin2017continual} rely on integrating previously saved or generated samples from earlier tasks with the current training data to facilitate learning new tasks. By replaying these past experiences, these techniques help the model to retain essential information from prior tasks, thereby effectively mitigating the problem of forgetting. 
{blue}{Dynamic architecture-based methods~\cite{mallya2018packnet,rosenfeld2018incremental,yoon2018lifelong,yan2021dynamically,wang2022foster,wang2022beef} involve dynamically extending or adapting the model's architecture to accommodate the demands of new tasks.} 
These approaches often adjust the model’s structure or its parameters, ensuring that the knowledge gained from previous tasks is protected while still enabling the model to learn new tasks effectively.
Regularization-based methods impose constraints on various aspects of the learning process, such as network weights~\cite{aljundi2018memory,kirkpatrick2017overcoming,zenke2017continual,10323204,chen2023skd}, intermediate features~\cite{hou2019learning}, or output probabilities~\cite{li2017learning,qiu2024incremental}. These constraints are designed to mitigate catastrophic forgetting by penalizing or restricting changes in the learned parameters that might otherwise disrupt previously acquired knowledge. In summary, each of these approaches offers a unique strategy for addressing the challenges of continual learning. Memory-based methods emphasize the importance of revisiting previous samples, dynamic architecture-based methods focus on adaptability and structural flexibility, while regularization-based methods prioritize stability and the careful management of learned parameters. Together, these methods contribute to the ongoing development of more robust and effective continual learning systems, capable of maintaining high levels of performance across a diverse range of tasks.

\subsection{NER}
\label{ner}

NER is a technique designed to identify specific words in unstructured texts that belong to predefined entity types \cite{li2020survey}. It serves as a vital component in more advanced applications such as information retrieval and question answering. This paper focuses primarily on flat NER, which aims to detect contiguous named entities that are not interrupted by non-entity words and do not overlap with other entities \cite{wang2022nested}. To achieve this, common sequence tagging schemes like ``BIO" (Beginning-Inside-Outside) are used to classify each token and group them together to form complete named entities.

Traditional NER approaches have primarily concentrated on developing various deep learning models to extract entities from unstructured text \cite{li2020survey,zhang2024ssr,zhang2024cross}. With the rise of deep learning, the BiLSTM architecture quickly became the go-to model for early NER tasks \cite{li2020survey}. In a BiLSTM, each word's representation is informed by the contextual information from both its preceding and succeeding words, making it highly effective for a range of tagging applications \cite{lample2016neural}.
As deep learning advanced, the introduction of PLMs like BERT~\cite{kenton2019bert}, RoBERTa~\cite{liu2019roberta}, and DeBERTa \cite{hedeberta} marked a significant shift in the NLP field. Transformer-based PLMs rapidly gained popularity due to their ability to learn rich, contextual embeddings from vast amounts of text. These models have set a new standard for NER tasks by providing superior input representations that are more nuanced and context-aware.

Despite these advancements, most current NER models are still designed to recognize a fixed set of predefined entity types. However, real-world applications require a more flexible approach, where NER models must continuously identify and adapt to new entity types without necessitating a complete retraining process. This dynamic capability is crucial for keeping pace with the evolving nature of language and ensuring that NER systems remain effective in diverse environments.

\subsection{CNER}
\label{cner}

To address the challenges posed by continually evolving entity types, CNER combines the continual learning paradigm with traditional NER models~\cite{monaikul2021continual,zhang2023decomposing,zhang2023rdp,zhang2023continual,yu2024flexible}. One notable approach, ExtendNER~\cite{monaikul2021continual}, investigates the use of KD in the context of CNER. In this approach, the new model learns to replicate the classifier probabilities generated by the old model, ensuring that the knowledge from earlier models is preserved. Another innovative method, L\&R~\cite{xia2022learn}, introduces a `learn-and-review' framework. The learning phase is similar to ExtendNER, but the reviewing phase goes further by generating synthetic samples of previously learned entity types to augment the current dataset, thereby enhancing the model's ability to retain old knowledge. CFNER~\cite{zheng2022distilling} presents a causal framework for CNER and distills causal effects from the non-entity type. 
Although these methods represent significant progress, they share a common limitation: they all focus on strictly distilling representations from the old model into the new one. This emphasis on preserving prior knowledge leads to models that exhibit high stability, but at the cost of reduced plasticity, meaning they struggle to adapt to new entity types effectively. This trade-off between stability and plasticity remains a critical challenge in the ongoing development of more flexible and adaptive CNER methods.

To address this limitation, our previous conference publication \cite{zhang2023continual} introduced a pooled features distillation loss to achieve a suitable balance between model stability and plasticity by consolidating representation dimensions properly. However, relying solely on distillation often results in limited improvements, as it only enforces consistency between the representations of old and new models. Therefore, this paper proposes a SPT method that balances old and new knowledge from both representation and weight perspectives, leading to improved CNER performance.

\section{Method}
\label{Method}

\subsection{Problem Formulation}
\label{problem}

CNER is designed to train a model through steps $t=1,..., T$, progressively learning an expanding set of entity types. At each step, the training set $\mathcal{D}_t$ includes pairs $(X^t, {Y}^t)$, where $X^t$ denotes an input token sequence of length $|X^t|$, and ${Y}^t$ is the corresponding ground truth label sequence in one-hot format. Importantly, ${Y}^t$ only contains labels for the current entity types $\mathcal{E}^t$; all other labels, including those for future entity types $\mathcal{E}^{t+1:T}$ and past entity types $\mathcal{E}^{1:t-1}$, are collapsed into the non-entity type $e_{o}$. At step $t$ ($t > 1$), given the previous model $\mathcal{M}_{t-1}$ (frozen) and the current training set $\mathcal{D}_t$, our goal is to update to a new model $\mathcal{M}_t$ that can recognize entities from all types encountered up to that point, represented by $\bigcup^{t}_{i=1}\mathcal{E}^i$.

\subsection{SPT}
\label{spt}

\begin{figure*}[t]
\centering 
  \includegraphics[width=1.0\linewidth]{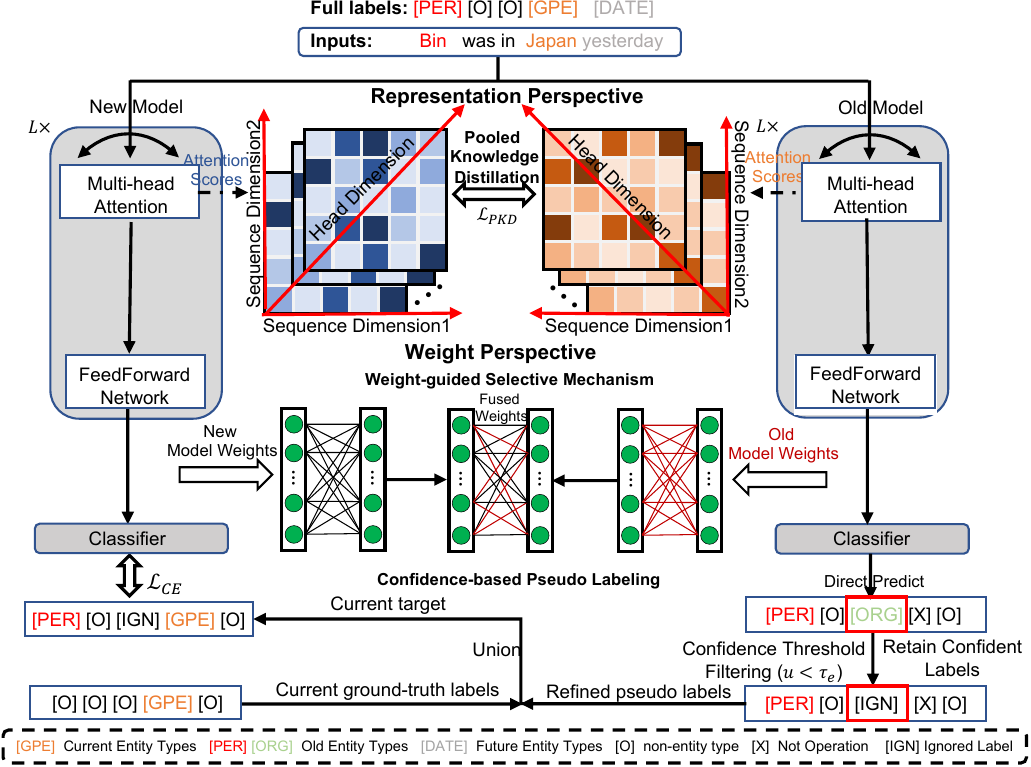}
\caption{Overview of our SPT method. It achieves a suitable balance between stability and plasticity from both representation and weight perspectives, aiming to effectively mitigate catastrophic forgetting. On the representation side, we enhance the original KD loss by integrating pooling operations, which help to consolidate the representation dimensions properly. Regarding weights, our method dynamically merges the weights of the old and new models. To prevent underfitting of new entity types, we introduce a weight-guided selective mechanism that selectively incorporates crucial weights associated with previous entity types from the old model into the new one. Additionally, we develop a confidence-based pseudo-labeling strategy to identify previous entity types within tokens currently labeled as the non-entity type, thereby effectively addressing the issue of semantic shift.}
\label{fig:model}
\end{figure*}

Figure~\ref{fig:model} provides an overview of our proposed SPT method for addressing the CNER problem. This method balances both old and new knowledge from the perspectives of representation (Section \ref{rep}) and weights (Section \ref{weight}). Additionally, to address the issue of semantic shifts in the non-entity type, we have designed a confidence-based pseudo-labeling strategy (Section \ref{pseudo}). Finally, Section \ref{overall} summarizes the overall optimization pipeline of our SPT method.

\subsubsection{Representation Perspective}
\label{rep}

Recent studies~\cite{clark2019does} indicate that attention scores learned by PLMs can embody extensive linguistic knowledge, encompassing both coreference and syntactical information, which are crucial for NER. In light of this, we first develop a KD loss intended to facilitate the transfer of linguistic knowledge embedded in attention scores from the previous model to the new one. This loss is formulated as follows:
 \begin{equation}
    \mathcal{L}_{\text{KD}} = \sum^{{K}}_{k=1} \sum^{{|X^t|}}_{i=1}\sum^{{|X^t|}}_{j=1} \Bigg|\Bigg|\bm{A}^{t}_{\ell,k,i,j} - \bm{A}^{t-1}_{\ell,k,i,j}\Bigg|\Bigg|^2
    \label{kd}\text{,}
 \end{equation}
 where $\bm{A}^t_{\ell}$ and $\bm{A}^{t-1}_{\ell} \in \mathbb{R}^{K \times |X^t| \times |X^t|}$ represent the attention scores for layer $\ell$ in models $\mathcal{M}_t$ and $\mathcal{M}_{t-1}$, respectively, with $\ell = 1, \ldots, L$ and $K$ denoting the number of attention heads.

  However, the KD loss described in Equation (\ref{kd}) can be seen as a rigid approach. This method forces $\mathcal{M}_t$ to precisely match the representations (\emph{i.e.}, attention scores) of $\mathcal{M}_{t-1}$. As a result, while it ensures stability (\emph{i.e.}, preservation of prior knowledge), it often hampers plasticity (\emph{i.e.}, the ability to incorporate new knowledge). Recent research supports the idea that the distillation process should strike a balance between stability and plasticity~\cite{DBLP:conf/eccv/DouillardCORV20,DBLP:conf/cvpr/DouillardCDC21,pelosin2022towards,kurmi2021not}.

  To achieve this, we introduce a pooling operation into the loss, which allows for greater plasticity by aggregating the pooled dimensions~\cite{DBLP:conf/eccv/DouillardCORV20,pelosin2022towards}. \textbf{The rationale behind this is that} minimal plasticity enforces an exact similarity between the old and new models, while increased plasticity relaxes this strict definition. Pooling reduces this similarity, and more extensive pooling provides a greater degree of flexibility.

By pooling the sequence dimensions, we obtain a more flexible loss that retains only the head dimension. This is formulated as follows:
 \begin{equation}
    \mathcal{L}_{\text{PKD-lax}} = \sum^{{K}}_{k=1} \Bigg|\Bigg|\sum^{{|X^t|}}_{i=1}\sum^{{|X^t|}}_{j=1}\bm{A}^{t}_{\ell,k,i,j} - \sum^{{|X^t|}}_{i=1}\sum^{{|X^t|}}_{j=1}\bm{A}^{t-1}_{\ell,k,i,j}\Bigg|\Bigg|^2
    \label{lax}\text{,}
 \end{equation}
 where we firstly perform pooling on the attention scores from both the new and old models along two sequence dimensions to obtain a one-dimensional tensor corresponding to the number of attention heads, and then we align them.

Our pooling-based framework enables the development of a more adaptable KD loss, achieving a better balance between plasticity and stability. According to Equation (\ref{lax}), we can moderately trade off plasticity to improve stability by applying less aggressive pooling. This is accomplished by aggregating statistics across either the head or sequence dimensions:
 \begin{equation}
 \begin{aligned}
    \mathcal{L}_{\text{PKD}} &= \sum^{{|X^t|}}_{i=1}\sum^{{|X^t|}}_{j=1} \Bigg|\Bigg|\sum^{{K}}_{k=1}\bm{A}^{t}_{\ell,k,i,j} - \sum^{{K}}_{k=1}\bm{A}^{t-1}_{\ell,k,i,j} \Bigg|\Bigg|^2\\
    &+ \sum^{{K}}_{k=1}\sum^{{|X^t|}}_{j=1} \Bigg|\Bigg|\sum^{{|X^t|}}_{i=1}\bm{A}^{t}_{\ell,k,i,j} - \sum^{{|X^t|}}_{i=1}\bm{A}^{t-1}_{\ell,k,i,j} \Bigg|\Bigg|^2\\
    & + \sum^{{K}}_{k=1}\sum^{{|X^t|}}_{i=1} \Bigg|\Bigg|\sum^{{|X^t|}}_{j=1}\bm{A}^{t}_{\ell,k,i,j} - \sum^{{|X^t|}}_{j=1}\bm{A}^{t-1}_{\ell,k,i,j} \Bigg|\Bigg|^2\\
 \end{aligned}
 \label{pkd}\text{,}
 \end{equation}
where we firstly perform pooling on the attention scores from both the new and old models separately along the attention head dimension and one of the sequence dimensions, resulting in three two-dimensional tensors, which are then aligned. $\mathcal{L}_{\text{PKD}}$ strikes a suitable balance between excessive rigidity, as indicated by Equation (\ref{kd}), and extreme leniency, as demonstrated in Equation (\ref{lax}).

\subsubsection{Weight Perspective}
\label{weight}

To further preserve old knowledge while enhancing the model's learning ability, we dynamically combine the weights of the old and new models without incurring additional computational costs. The formulation is as follows:
\begin{equation}
    \bm{\theta}^{t}_{\text{balanced}} = \alpha \bm{\theta}^{t-1} + (1-\alpha)\bm{\theta}^{t}\text{,}\\
    \label{wf_eq}
\end{equation}
where $\bm{\theta}^{t-1}$ represents the final model from the previous step $t$-$1$, considered the optimal repository of old knowledge, $\bm{\theta}^{t}$ denotes the model trained at the current step $t$, which includes discriminative information for new entity types, and $\alpha$ is the dynamic balance factor, associated with the ratio of the number of entity types learned at step $t$ to the total number of entity types encountered over time, computed as follows:
\begin{equation}
    \alpha = 1 - \sqrt{\frac{E^t}{\sum_{m=1}^t E^m +1}}\text{,}
    \label{dynamic}
\end{equation}
where $E^m$ represents the number of entity types at step $m$.

However, Equation (\ref{wf_eq}) merges the weights of the old and new models indiscriminately, which can introduce bias towards the old model’s weights and result in underfitting of new entity types. To address this, we propose a weight-guided selective mechanism that merges the old and new model weights based on their importance for the old entity types. This method effectively learns new entity types while preserving previously learned knowledge. \textbf{The motivation is that} instead of merging all old weights indiscriminately, it is more effective to combine only those old weights that are crucial for the old entity types with their corresponding new weights. This approach minimizes the interference from less important old weights in the merging process.

Specifically, the weight importance $F^{t-1}$ is determined by the Fisher information of the corresponding gradients at step $t$-$1$ \cite{pascanu2013revisiting,cong2024cs2k}. After learning at step $t$, the fusion of old and new model weights based on the weight importance $F^{t-1}$ for old entity types is formulated as follows:
\begin{equation}
\footnotesize
\bm{\theta}^{t}_{\text{balanced},i} = \left\{
\begin{aligned}
  \alpha \bm{\theta}^{t-1}_i + (1-\alpha)\bm{\theta}^{t}_i \ \ \ &\text{If}\ \ F^{t-1}_i > \text{TopK}(F^{t-1}, \gamma|\theta^{t-1}|)  \\
  \bm{\theta}^{t}_i \ \ \ &\text{Otherwise,}
\end{aligned}
\right.
\label{fused}
\end{equation}
where $F^{t-1}_i$ represents the importance of the $i$-th old model weight $\theta^{t-1}_i$ for old entity types, $|\theta^{t-1}|$ denotes the total number of weights in the old model $\mathcal{M}_{t-1}$, \text{TopK}($F^{t-1}$, $\gamma|\theta^{t-1}|$) indicates the $\gamma|\theta^{t-1}|$-th largest value in $F^{t-1}$, and $\gamma$ determines how many important old weights are selected.

Similar to $\alpha$ in Equation (\ref{dynamic}), $\gamma$ is a dynamic factor that can be automatically adjusted across different steps and settings. Specifically, $\gamma$ acts as a threshold for distinguishing weight importance and is closely related to the difference in the number of entity types acquired at step $t$ compared to those learned previously. Therefore, the calculation of $\gamma$ is designed as follows:
\begin{equation}
    \gamma = (1+e^{\frac{E^t-\sum_{m=1}^{t-1}E^m-1}{\sum_{m=1}^tE^m+1}})^{-1}\text{.}
\end{equation}

In conclusion, our weight-guided selective mechanism identifies crucial model weights for old entity types to mitigate forgetting, while simultaneously recognizing new entity types by retaining the remaining weights from the new model.

\subsubsection{Confidence-based Pseudo-labeling}
\label{pseudo}

\begin{figure}[t]
\centering
  \includegraphics[width=1.0\linewidth]{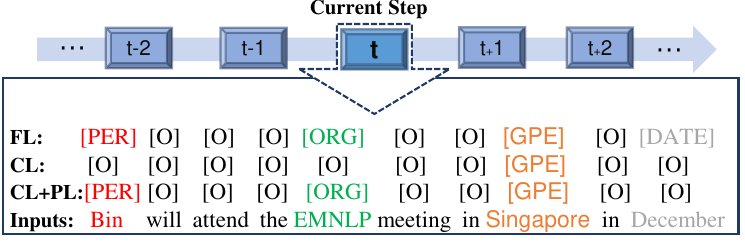} 
\caption{Example illustrating the semantic shift of the non-entity type. \textbf{FL}, \textbf{CL}, and \textbf{PL} represent Full ground-truth Labels, Current ground-truth Labels, and Pseudo Labels, respectively. At the current step $t$, old entity types (\emph{e.g.}, \textcolor{green}{[ORG]} (\texttt{Organization}) and \textcolor{red}{[PER]} (\texttt{Person})) and future entity types (\emph{e.g.}, \textcolor{gray}{[DATE]} (\texttt{Date})) are masked as \textcolor{black}{[O]} (\texttt{the non-entity type}), while \textcolor{orange}{[GPE]} (\texttt{Countries}) is the current entity type being learned. This masking leads to the semantic shift issue for the non-entity type (illustrated in the second row \textbf{CL}). By using PL from the previous model to augment the CL as the current classification target, the problem of semantic shift can be effectively mitigated (shown in the third row \textbf{CL+PL}).}
\label{fig:pseudo}
\end{figure}

As previously noted, the current ground-truth label sequence $Y^t$ includes labels solely for the current entity types $\mathcal{E}^t$. All other labels—such as potential old entity types $\mathcal{E}^{1:t-1}$ or future entity types $\mathcal{E}^{t+1:T}$—are collapsed into the non-entity type $e_{o}$. This results in a semantic shift phenomenon for the non-entity type \cite{zhang2023continual}. As illustrated in Figure~\ref{fig:pseudo} (second row \textbf{CL}), old entity types \textcolor{green}{[ORG]} (\texttt{Organization}) and \textcolor{red}{[PER]} (\texttt{Person}), learned in prior steps (\emph{e.g.}, $t$-$1$, $t$-$2$), along with the future entity type \textcolor{gray}{[DATE]} (\texttt{Date}), which will be learned in upcoming steps (\emph{e.g.}, $t$+$1$, $t$+$2$), are all marked as the non-entity type at the current step $t$. Here, \textcolor{orange}{[GPE]} (\texttt{Countries}) is the current entity type to be learned at step $t$. The lack of mechanisms to differentiate between tokens related to old entity types and the genuine non-entity type may worsen the issue of catastrophic forgetting.

To address this, we employ a confidence-based pseudo-labeling strategy (proposed in our previous conference publication \cite{zhang2023continual}) to augment the current ground-truth labels by integrating predictions from the previous model (illustrated in the third row \textbf{CL+PL} of Figure \ref{fig:pseudo}). This approach generates more informative labels for classification. 
Formally, we denote the cardinality of current entity types as $E^t = \text{card}(\mathcal{E}^t)$. The predictions of the current model $\mathcal{M}_t$ are represented by $\hat{Y}^t \in \mathbb{R}^{|X^t| \times (1 + E^1 + \ldots + E^t)}$. We define $\tilde{Y}^t \in \mathbb{R}^{|X^t| \times (1 + E^1 + \ldots + E^t)}$ as the current target, computed using the one-hot ground-truth label sequence ${Y}^t \in \mathbb{R}^{|X^t| \times (1 + E^1 + \ldots + E^t)}$ and the pseudo-labels derived from the predictions $\hat{Y}^{t-1} \in \mathbb{R}^{|X^t| \times (1 + E^1 + \ldots + E^{t-1})}$ of the previous model $\mathcal{M}_{t-1}$. The process is described as follows:
\begin{equation}
\footnotesize
\bm{\widetilde{Y}}^t_{i,e}= \mkern-5mu \left\{\begin{array}{ll}
\mkern-10mu 1 \mkern-27mu & \ \ \ \text {if}\ \ \bm{Y}^{t}_{i,e_{o}}=0\ \ \&\ \ e = \argmax \limits_{e' \in \mathcal{E}^{t}}\ \bm{Y}^{t}_{i,e'} \\
\mkern-10mu 1 \mkern-27mu & \ \ \ \text {if}\ \ \bm{Y}^{t}_{i,e_{o}}=1\ \ \&\ \ e = \mkern-8mu \argmax \limits_{e' \in e_o\cup\mathcal{E}^{1:t-1}} \mkern-6mu\ \ \bm{\widehat{Y}}^{t-1}_{i,e'}\ \&\ u<\tau_e \\
\mkern-10mu 0 \mkern-27mu & \ \ \ \text {otherwise}\\
\end{array}\right.
\label{eq:pseudo_better}\text{,}
\end{equation}
in other words, if a token is not labeled as the non-entity type $e_{o}$, we use the ground-truth labels (as shown in the first row of Equation (\ref{eq:pseudo_better})). If it is labeled as the non-entity type $e_{o}$, we instead use the pseudo-labels predicted by the old model (as indicated in the second row of Equation (\ref{eq:pseudo_better})). To minimize prediction errors from the old model, this confidence-based pseudo-labeling strategy employs entropy as a measure of uncertainty, using the median entropy as the confidence threshold. Specifically, before starting the learning process for step $t$, we first feed the current dataset $\mathcal{D}_t$ into the old model $\mathcal{M}_{t-1}$ for inference and compute the entropy $u$ of the output probability distribution for each non-entity token. Next, for each non-entity token, we categorize them based on their predicted old type, determined by the index of the highest output probability predicted by the previous model $\mathcal{M}_{t-1}$. This classification allows us to calculate the median entropy $\tau_e$ for each old type $e \in e_o \cup \mathcal{E}^{1:t-1}$ within the respective groups. Pseudo-labels are retained only when the old model is sufficiently ``confident'' ($u < \alpha_e$).

After applying pseudo-labeling, the Cross-Entropy (CE) loss can be expressed as follows:
\begin{equation}
    \mathcal{L}_{\text{CE}}=- \frac{1}{|X^t|} \sum_{i}^{|X^t|} \eta_i \bm{\widetilde{Y}}^t_{i}\log \bm{\widehat{Y}}^t_{i}\text{,}
\label{eq:pseudo_loss}
\end{equation}
where $\eta_i$ represents the weight of the token at position $i$ in the sequence $X^t$. This weight adaptively assigns different values to tokens of various types based on their quantity, and is computed as follows:
\begin{equation}
\eta_i=\left\{
\begin{aligned}
&0.5 + \sigma(\frac{N^{\text{old}}}{N^{\text{new}}})\ \ \ \text{If}\ \ \bm{\widetilde{Y}}^t_{i}\in \mathcal{E}^{1:t-1} \\
&1.0\ \ \ \ \ \ \ \ \ \ \ \ \ \ \ \ \ \ \ \ \ \ \ \text{Otherwise,}
\end{aligned}
\right.
\end{equation}
where $N^{\text{old}}$, $N^{\text{new}}$, and $\sigma(\cdot)$ denote the number of tokens associated with old entity types $\mathcal{E}^{1:t-1}$, the number of tokens corresponding to new entity types $\mathcal{E}^t$, and the sigmoid function, respectively.

\begin{algorithm}[t]
    \caption{The optimization process for our SPT method.}
    \label{algo1}
    \begin{algorithmic}
        \Require Initial weights $\theta^{0}$, total steps $T$, current dataset $\mathcal{D}_{t}$
        \State $t \gets 1$ \textcolor{blue}{\Comment{Continual training step}}
        \While{$t \le T$}
        \State{$\theta^{t} \gets \theta^{t-1}$}
        
        \State{$j \gets 1$} \textcolor{blue}{\Comment{Iteration step, Train on the current training set}}
        \While{not converged}
        \State{Sample mini-batch $\{ \bm{X}^t, \bm{Y}^t \} \sim \mathcal{D}_t$}
        \State{$\theta^{t}_{j+1} \gets \theta^{t}_j-{\text{lr}} \nabla_{\mathcal{L}_{\text{Total}}\text{ from Equation (\ref{eq:complete_loss})}}$}
        \State{$j \gets j+1$}
        \EndWhile
        \State{{Get the fused weight $\theta^t_{\text{balanced}}$ using Equation (\ref{fused})}} \textcolor{blue}{\Comment{Test on all entity types seen so far}}
        \State{$\theta^{t} \gets \theta^t_{\text{balanced}}$}
        \State{$t \gets t+1$}
        \EndWhile
    \end{algorithmic}
\end{algorithm}

\subsubsection{Overall Optimization Pipeline}
\label{overall}

Finally, the comprehensive optimization objective for our SPT method is:
\begin{equation}
    \mathcal{L}_{\text{Total}}(\Theta_t) = \underbrace{\strut \mathcal{L}_{\text{CE}}}_\text{Classification} + \lambda\underbrace{\strut \mathcal{L}_{\text{PKD}}}_\text{Distillation}\text{,}
\label{eq:complete_loss}
\end{equation}
where $\lambda$ is a hyperparameter to balance the losses, and $\Theta_t$ represents the set of learnable parameters for the model $\mathcal{M}_t$. 
The optimization process is summarized in Algorithm \ref{algo1}.

\section{Experimental Configuration}
\label{ex_details}

To ensure a fair comparison with SOTA CNER methods, we adhere to the experimental configuration detailed in CFNER~\cite{zheng2022distilling} and CPFD~\cite{zhang2023continual}. This approach involves using the same benchmark datasets (Section \ref{dataset}), CNER settings (Section \ref{cner_setting}), evaluation metrics (Section \ref{metrics}), competing baselines (Section \ref{baseline}), and core implementation specifics (Section \ref{details}).

\subsection{Benchmark Datasets}
\label{dataset}

We evaluate our SPT method using three widely used NER datasets: CoNLL2003~\cite{sang1837introduction}, I2B2~\cite{murphy2010serving}, and OntoNotes5~\cite{hovy2006ontonotes}. A summary of the dataset statistics is provided in Table~\ref{tab:dataset_statistics}.

To divide the training set into distinct slices that represent different continual learning steps, we employ the greedy sampling algorithm \cite{zheng2022distilling}. This technique ensures that samples for each entity type are primarily concentrated within their respective slices, thus more accurately reflecting real-world scenarios. Within each slice, only labels relevant to the currently learned entity types are retained, while all others are categorized as the non-entity type. More details on this greedy sampling algorithm can be found in CFNER's Appendix B \cite{zheng2022distilling}.

\begin{table}[t]
  \centering
  \caption{The summary for three widely used NER datasets.}
  \resizebox{1.0\linewidth}{!}{
    \begin{tabular}{ccccc}
    \toprule
     \textbf{Dataset}     & \multicolumn{1}{l}{\# \textbf{Entity Type}} & \# \textbf{Sample}  & \multicolumn{2}{c}{\textbf{Entity Type Sequence (Alphabetical Order)}} \\
    \midrule
    CoNLL2003~\cite{sang1837introduction} & 4     & 21k    & \multicolumn{2}{l}{\begin{tabular}[1]{l}LOCATION, MISC, ORGANISATION, PERSON\end{tabular}} \\
    \midrule
    I2B2 \cite{murphy2010serving}  & 16    & 141k   & \multicolumn{2}{c}{\begin{tabular}[1]{l}AGE, CITY, COUNTRY, DATE, DOCTOR, HOSPITAL, \\  IDNUM, MEDICALRECORD, ORGANIZATION, \\PATIENT, PHONE, PROFESSION, STATE, STREET, \\USERNAME, ZIP\end{tabular}} \\
    \midrule
    OntoNotes5 \cite{hovy2006ontonotes} & 18    & 77k   & \multicolumn{2}{c}{\begin{tabular}[1]{l}CARDINAL, DATE, EVENT, FAC, GPE, LANGUAGE,\\ LAW, LOC, MONEY, NORP, ORDINAL, ORG,\\ PERCENT, PERSON, PRODUCT, QUANTITY, TIME,\\ WORK\_OF\_ART\end{tabular}} \\
    \bottomrule
    \end{tabular}%
    }
  \label{tab:dataset_statistics}%
\end{table}%

\subsection{CNER Settings}
\label{cner_setting}
In accordance with CFNER~\cite{zheng2022distilling} and CPFD~\cite{zhang2023continual}, we investigate two distinct CNER scenarios for each dataset. In the first scenario, the base model (initial step) is trained with the same number of entity types as in subsequent steps. In the second scenario, the base model is trained with half of the total entity types. The first scenario is more challenging, whereas the second better simulates real-world conditions by allowing the model to acquire sufficient knowledge before continual learning begins.
Entity types are learned sequentially during training, specifically in alphabetical order, with each data slice used to train the models in sequence. The base model is trained with FG entity types, while each continual learning step introduces PG entity types, denoted as FG-a-PG-b. For the CoNLL2003 dataset~\cite{sang1837introduction}, we use two settings: FG-1-PG-1 and FG-2-PG-1. For the I2B2~\cite{murphy2010serving} and OntoNotes5~\cite{hovy2006ontonotes} datasets, four settings are employed: FG-1-PG-1, FG-2-PG-2, FG-8-PG-1, and FG-8-PG-2.
During evaluation, only the labels relevant to the current entity types are retained in the validation set, with all other labels marked as the non-entity type. For each continual learning step, the model with the best validation performance is selected for testing and for the next continual learning phase. In the testing phase, labels for all previously learned entity types are preserved in the test set, while others are categorized as the non-entity type.

\subsection{Evaluation Metrics}
\label{metrics}
At each continual learning step, we first compute the F1 score for each entity type. We then calculate the macro-average F1 scores (Ma-F1) for old entity types, new entity types, and all entity types up to the current step. Moreover, we compute the micro-average F1 score (Mi-F1) for all current entity type. Average scores are reported across all steps, including the initial step (except for old and new entity types). We also present line plots to show the changes in scores over the steps.

\subsection{Competing Baselines}
\label{baseline}
We include several baselines for comparison, including SOTA CNER methods: ExtendNER~\cite{monaikul2021continual}, CFNER~\cite{zheng2022distilling}, IS3 \cite{qiu2024incremental}, and CPFD~\cite{zhang2023continual}, the latter of which was introduced in our prior conference publication. Moreover, we evaluate the lower-bound method, Finetuning (FT), which directly finetunes the model on new data without applying any anti-forgetting techniques.
Furthermore, we incorporate continual learning approaches from computer vision, such as PODNet~\cite{DBLP:conf/eccv/DouillardCORV20}, LUCIR~\cite{DBLP:conf/cvpr/HouPLWL19}, and Self-Training (ST)~\cite{DBLP:journals/corr/abs-1909-08383}, adapted to the CNER context~\cite{zheng2022distilling}. Detailed descriptions of these baselines are provided below:

\begin{table*}[t]
\centering
\caption{Comparisons with baselines on the CoNLL2003~\cite{sang1837introduction} dataset. The \textcolor{deepred}{\textbf{red}} indicates the highest result, while the \textcolor{blue}{\textbf{blue}} signifies the second highest result. $*$ indicates results obtained from our re-implementation based on publicly available code. All other baseline results are directly cited from CPFD~\cite{zhang2023continual}.}

\resizebox{1.0\linewidth}{!}{%
\begin{tabular}{@{}ccccccccc@{}}
\toprule
&\multicolumn{4}{c}{FG-1-PG-1} & \multicolumn{4}{c}{FG-2-PG-1}  \\ \cmidrule(l){2-5} \cmidrule(l){6-9} 
&\multicolumn{1}{c}{Old Entity Types} & \multicolumn{1}{c}{New Entity Types}&\multicolumn{2}{c}{All Entity Types}& \multicolumn{1}{c}{Old Entity Types} & \multicolumn{1}{c}{New Entity Types}&\multicolumn{2}{c}{All Entity Types} \\
\cmidrule(l){4-5} \cmidrule(l){8-9}
 \multirow{-4}{*}{Baselines} & Avg. Ma-F1   & Avg. Ma-F1  & Avg. Mi-F1  & Avg. Ma-F1  & Avg. Ma-F1   & Avg. Ma-F1  &Avg. Mi-F1  &Avg. Ma-F1  \\ \midrule
 
  FT &-- &-- & 50.84±0.10 & 40.64±0.16 &-- &-- & 57.45±0.05 & 43.58±0.18 \\

{$\text{FT}^{*}$} &0.00±0.00 &74.42±0.35 & 50.62±0.08 & 40.34±0.08 &0.14±0.06 &80.81±0.36 & 57.38±0.12 & 43.10±0.05 \\
 
 PODNet~\cite{DBLP:conf/eccv/DouillardCORV20} &-- &-- & 36.74±0.52 & 29.43±0.28 &-- &-- & 59.12±0.54 & 58.39±0.99 \\

{$\text{PODNet}^{*}$}~\cite{DBLP:conf/eccv/DouillardCORV20} &11.67±0.56 &39.41±0.48 & 43.60±0.30 & 36.74±0.30 &32.11±3.69 &54.51±3.75 & 60.43±1.62 & 53.19±2.05 \\

 LUCIR~\cite{DBLP:conf/cvpr/HouPLWL19} &-- &-- & 74.15±0.43 & 70.48±0.66 &-- &-- & 80.53±0.31 & 77.33±0.31 \\

{$\text{LUCIR}^{*}$}~\cite{DBLP:conf/cvpr/HouPLWL19} &73.19±0.81 &72.42±0.23 & 77.10±0.13 & 74.21±0.41 &64.91±0.96 &83.04±0.37 & 80.59±0.33 & 74.08±0.53 \\

 ST~\cite{DBLP:conf/wacv/RosenbergHS05} & --& --& 76.17±0.91 & 72.88±1.12 &-- & --& 76.65±0.24 & 66.72±0.11 \\

{$\text{ST}^{*}$}~\cite{DBLP:conf/wacv/RosenbergHS05} & 67.09±0.68& 68.95±1.25& 73.71±0.72 & 70.15±0.59 &49.20±1.58 & 77.37±1.34& 76.07±0.18 & 65.72±0.67 \\

 ExtendNER~\cite{monaikul2021continual} &-- & --& 76.36±0.98 & 73.04±1.80 &-- &-- & 76.66±0.66 & 66.36±0.64 \\

  $\text{ExtendNER}^{*}$~\cite{monaikul2021continual} & 70.42±0.22 & 71.84±0.69 & 76.07±0.35 & 73.06±0.29 & 56.33±2.00 & 79.34±0.71 & 77.89±0.42 & 69.92±1.02 \\

 CFNER~\cite{zheng2022distilling} &-- &-- & 80.91±0.29 & 79.11±0.50 &-- &-- & 80.83±0.36 & 75.20±0.32  \\

  $\text{CFNER}^{*}$~\cite{zheng2022distilling} & 76.82±0.58 & 78.87±0.44 & 80.29±0.21 & 78.44±0.24 & 69.94±1.63 & 84.08±0.26 & 81.52±0.43 & 77.20±0.82  \\

  CPFD \cite{zhang2023continual}  &-- &-- & \textcolor{blue}{\textbf{82.24±0.63}} & \textcolor{blue}{\textbf{79.94±0.66}} &-- &-- & 85.70±0.19 & \textcolor{blue}{\textbf{83.49±0.16}}  \\

 $\text{CPFD}^{*}$~\cite{zhang2023continual} & \textcolor{blue}{\textbf{78.30±0.67}} & \textcolor{blue}{\textbf{81.42±0.49}} & 81.91±0.17 & 79.48±0.26 & \textcolor{blue}{\textbf{81.29±1.11}} & \textcolor{blue}{\textbf{88.93±0.42}} & \textcolor{blue}{\textbf{85.76±0.36}} & 83.17±0.50  \\

 \midrule

 \textbf{SPT (Ours)} & \textcolor{deepred}{\textbf{80.52±0.21}} & \textcolor{deepred}{\textbf{83.57±0.26}} & \textcolor{deepred}{\textbf{84.22±0.41}} & \textcolor{deepred}{\textbf{81.45±0.19}} & \textcolor{deepred}{\textbf{82.88±0.76}} & \textcolor{deepred}{\textbf{90.20±0.52}} & \textcolor{deepred}{\textbf{87.22±0.58}} & \textcolor{deepred}{\textbf{84.99±0.73}} \\

\textbf{Imp.} & $\Uparrow$\textbf{2.22} & $\Uparrow$\textbf{2.15}& $\Uparrow$\textbf{1.98}& $\Uparrow$\textbf{1.51}& $\Uparrow$\textbf{1.59}& $\Uparrow$\textbf{1.27}& $\Uparrow$\textbf{1.46}& $\Uparrow$\textbf{1.50} \\

\bottomrule
\end{tabular}
}
\label{tab:main_result_1_full}
\end{table*}

\begin{table*}[t]
\centering
\caption{Comparisons with baselines on the I2B2~\cite{murphy2010serving} dataset. The \textcolor{deepred}{\textbf{red}} indicates the highest result, while the \textcolor{blue}{\textbf{blue}} signifies the second highest result. $*$ indicates results obtained from our re-implementation based on publicly available code. All other baseline results are directly cited from CPFD~\cite{zhang2023continual} or their original papers.}

\resizebox{1.0\linewidth}{!}{%
\begin{tabular}{@{}ccccccccc@{}}
\toprule
&\multicolumn{4}{c}{FG-1-PG-1} & \multicolumn{4}{c}{FG-2-PG-2}  \\ \cmidrule(l){2-5} \cmidrule(l){6-9} 
&\multicolumn{1}{c}{Old Entity Types} & \multicolumn{1}{c}{New Entity Types}&\multicolumn{2}{c}{All Entity Types}& \multicolumn{1}{c}{Old Entity Types} & \multicolumn{1}{c}{New Entity Types}&\multicolumn{2}{c}{All Entity Types} \\
\cmidrule(l){4-5} \cmidrule(l){8-9}
 \multirow{-4}{*}{Baselines} & Avg. Ma-F1   & Avg. Ma-F1  & Avg. Mi-F1  & Avg. Ma-F1  & Avg. Ma-F1   & Avg. Ma-F1  &Avg. Mi-F1  &Avg. Ma-F1  \\\midrule
 
 FT &-- &-- & 17.43±0.54 & 13.81±1.14 & --&-- & 28.57±0.26 & 21.43±0.41 \\

{$\text{FT}^{*}$} &3.18±0.35 &44.58±0.85 & 17.84±0.15 & 14.49±0.31 &1.94±0.20&57.13±0.55 & 28.50±0.13 & 21.40±0.08 \\
 
 PODNet~\cite{DBLP:conf/eccv/DouillardCORV20} &-- &-- & 12.31±0.35 & 17.14±1.03 &-- &-- & 34.67±2.65 & 24.62±1.76 \\

{$\text{PODNet}^{*}$}~\cite{DBLP:conf/eccv/DouillardCORV20} &18.36±1.59 &6.88±0.63 & 13.88±1.26 & 18.83±1.34 &18.72±1.18 &13.54±2.20 & 25.57±1.15 & 22.40±1.14 \\

 LUCIR~\cite{DBLP:conf/cvpr/HouPLWL19} &-- &-- & 43.86±2.43 & 31.31±1.62 &-- & --& 64.32±0.76 & 43.53±0.59 \\

{$\text{LUCIR}^{*}$}~\cite{DBLP:conf/cvpr/HouPLWL19} &40.66±0.58 &39.99±0.23 & 63.47±0.55 & 41.95±0.44 &44.69±0.28 & 52.26±0.47& 71.21±0.53 & 48.42±0.28 \\

 ST~\cite{DBLP:conf/wacv/RosenbergHS05} &-- &-- & 31.98±2.12 & 14.76±1.31 &-- &-- & 55.44±4.78 & 33.38±3.13\\

{$\text{ST}^{*}$}~\cite{DBLP:conf/wacv/RosenbergHS05} &13.13±7.10 &25.81±13.40 & 39.75±17.50 & 20.11±7.76 &23.48±6.60 &46.49±2.27 & 52.52±16.38 & 33.60±4.96\\

 ExtendNER~\cite{monaikul2021continual} &-- &-- & 42.85±2.86 & 24.05±1.35 &-- &-- & 57.01±4.14 & 35.29±3.38 \\

   $\text{ExtendNER}^{*}$~\cite{monaikul2021continual} & 15.67±2.47 & 30.01±5.07 & 41.65±10.11 & 23.11±2.70 & 35.66±2.26 & 50.78±1.17 & 67.60±1.15 & 42.58±1.59 \\

 CFNER~\cite{zheng2022distilling} &-- &-- & 62.73±3.62 & 36.26±2.24 &-- & --& 71.98±0.50 & 49.09±1.38  \\

 $\text{CFNER}^{*}$~\cite{zheng2022distilling} & 34.35±1.00 & 45.70±1.23 & 64.79±0.26 & 37.79±0.65 & 48.14±1.10 & 56.56±1.08 & 72.58±0.59 & 51.71±0.84  \\

{IS3}~\cite{qiu2024incremental} &-- &-- & -- & \textcolor{blue}{\textbf{56.87±0.56}} &-- & --& -- & \textcolor{blue}{\textbf{61.83±0.87}}  \\

 CPFD~\cite{zhang2023continual} &-- &-- & \textcolor{blue}{\textbf{74.19±0.95}} & 48.34±1.45 &-- & --& \textcolor{blue}{\textbf{78.19±0.58}} & 56.04±1.22  \\

 $\text{CPFD}^{*}$~\cite{zhang2023continual} & \textcolor{blue}{\textbf{47.01±0.84}} & \textcolor{blue}{\textbf{54.91±1.09}} & 74.01±0.82 & 49.09±0.58 & \textcolor{blue}{\textbf{49.83±1.58}} & \textcolor{blue}{\textbf{62.76±0.61}} & 77.83±0.39 & 53.95±1.11 \\

 \midrule

 \textbf{SPT (Ours)} & \textcolor{deepred}{\textbf{61.91±2.12}} & \textcolor{deepred}{\textbf{61.45±0.95}} & \textcolor{deepred}{\textbf{77.81±1.17}} & \textcolor{deepred}{\textbf{62.32±1.88}} & \textcolor{deepred}{\textbf{64.58±0.55}} & \textcolor{deepred}{\textbf{66.84±0.28}} & \textcolor{deepred}{\textbf{81.06±0.12}} & \textcolor{deepred}{\textbf{65.04±0.36}} \\
 
\textbf{Imp.} & $\Uparrow$\textbf{14.90}& $\Uparrow$\textbf{6.54}& $\Uparrow$\textbf{3.62}& $\Uparrow$\textbf{5.45}& $\Uparrow$\textbf{14.75}& $\Uparrow$\textbf{4.08}& $\Uparrow$\textbf{2.87}& $\Uparrow$\textbf{3.21} \\

\midrule

&\multicolumn{4}{c}{FG-8-PG-1} & \multicolumn{4}{c}{FG-8-PG-2}  \\ \cmidrule(l){2-5} \cmidrule(l){6-9} 
&\multicolumn{1}{c}{Old Entity Types} & \multicolumn{1}{c}{New Entity Types}&\multicolumn{2}{c}{All Entity Types}& \multicolumn{1}{c}{Old Entity Types} & \multicolumn{1}{c}{New Entity Types}&\multicolumn{2}{c}{All Entity Types} \\
\cmidrule(l){4-5} \cmidrule(l){8-9}
 \multirow{-4}{*}{Baselines} & Avg. Ma-F1   & Avg. Ma-F1  & Avg. Mi-F1  & Avg. Ma-F1  & Avg. Ma-F1   & Avg. Ma-F1  &Avg. Mi-F1  &Avg. Ma-F1  \\\midrule
 
 FT &-- &-- & 20.83±1.78 & 18.11±1.66 &-- &-- & 23.60±0.15 & 23.54±0.38 \\

{$\text{FT}^{*}$} &9.89±0.49 &33.64±1.03 & 21.61±0.27 & 19.41±0.41 &4.50±0.59 &42.36±0.49 & 24.39±0.30 & 24.51±0.36 \\
 
 PODNet~\cite{DBLP:conf/eccv/DouillardCORV20} &-- &-- & 39.26±1.38 & 27.23±0.93 &-- &-- & 36.22±12.9 & 26.08±7.42 \\

{$\text{PODNet}^{*}$}~\cite{DBLP:conf/eccv/DouillardCORV20} &21.10±1.12 &8.64±0.61 & 38.73±2.33 & 26.71±0.92 &21.23±0.20 &11.41±0.71 & 47.81±0.47 & 31.80±0.19 \\

 LUCIR~\cite{DBLP:conf/cvpr/HouPLWL19} &-- &-- & 57.86±0.87& 33.04±0.39 &-- &-- & 68.54±0.27 & 46.94±0.63 \\

{$\text{LUCIR}^{*}$}~\cite{DBLP:conf/cvpr/HouPLWL19} &31.04±0.57 &31.07±0.87 & 59.54±1.58& 36.46±0.47 &42.14±0.67 &41.50±0.50 & 69.02±0.44 & 49.59±0.50 \\

 ST~\cite{DBLP:conf/wacv/RosenbergHS05} &-- &-- & 49.51±1.35 & 23.77±1.01 &-- &-- & 48.94±6.78 & 29.00±3.04\\

{$\text{ST}^{*}$}~\cite{DBLP:conf/wacv/RosenbergHS05} &19.02±1.02 &28.37±1.57 & 44.90±3.09 & 26.53±0.91 &25.43±1.83 &39.25±0.60 & 58.80±2.61 & 38.06±1.24\\

 ExtendNER~\cite{monaikul2021continual} &-- &-- & 43.95±2.01 & 23.12±1.79 &-- &-- & 52.25±5.36 & 30.93±2.77 \\

   $\text{ExtendNER}^{*}$~\cite{monaikul2021continual} & 20.33±0.99 & 27.65±1.30 & 45.14±2.91 & 27.41±0.88 & 27.01±2.10 & 39.22±1.19 & 56.48±2.41 & 38.88±1.38 \\

 CFNER~\cite{zheng2022distilling} &-- &-- & 59.79±1.70 & 37.30±1.15 &-- &-- & 69.07±0.89 & 51.09±1.05 \\
 
 $\text{CFNER}^{*}$~\cite{zheng2022distilling} & 31.13±1.56 & 37.95±2.08 & 56.66±3.22 & 36.84±1.35 & 43.94±1.25 & 49.93±1.19 & 69.12±0.94 & 51.61±0.87 \\

{IS3}~\cite{qiu2024incremental} &-- &-- & -- & \textcolor{blue}{\textbf{58.38±1.35}} &-- & --& -- & 63.03±1.07  \\

 CPFD~\cite{zhang2023continual} &-- &-- & \textcolor{blue}{\textbf{74.75±1.35}} & 56.19±2.46 &-- &-- & \textcolor{blue}{\textbf{81.05±0.87}} & \textcolor{blue}{\textbf{65.04±1.13}} \\
 
 $\text{CPFD}^{*}$~\cite{zhang2023continual} & \textcolor{blue}{\textbf{52.17±1.59}} & \textcolor{blue}{\textbf{51.22±2.72}} & 73.80±1.95 & 55.06±1.33 & \textcolor{blue}{\textbf{59.70±2.03}} & \textcolor{blue}{\textbf{56.18±1.61}} & 80.20±0.79 & 63.25±1.51 \\

 \midrule

 \textbf{SPT (Ours)} & \textcolor{deepred}{\textbf{59.74±0.95}} & \textcolor{deepred}{\textbf{55.16±2.17}} & \textcolor{deepred}{\textbf{75.81±0.42}} & \textcolor{deepred}{\textbf{60.44±0.95}} & \textcolor{deepred}{\textbf{68.86±1.05}} & \textcolor{deepred}{\textbf{61.45±0.98}} & \textcolor{deepred}{\textbf{84.63±0.31}} & \textcolor{deepred}{\textbf{70.13±0.73}} \\
 
\textbf{Improve} & $\Uparrow$\textbf{7.57}& $\Uparrow$\textbf{3.94}& $\Uparrow$\textbf{1.06}& $\Uparrow$\textbf{2.06}& $\Uparrow$\textbf{9.16}& $\Uparrow$\textbf{5.27}& $\Uparrow$\textbf{3.58}& $\Uparrow$\textbf{5.09} \\

\bottomrule
\end{tabular}
}
\label{tab:main_result_2_full}
\end{table*}

\begin{itemize}

\item PODNet \cite{DBLP:conf/eccv/DouillardCORV20}: Designed to address catastrophic forgetting in continual learning for image classification, PODNet has been adapted for the CNER scenario \cite{zheng2022distilling}. Its total loss includes both classification and distillation losses. For classification, PODNet employs neighborhood component analysis loss rather than the traditional cross-entropy loss. For distillation, it applies constraints to the output of each intermediate layer.

\item LUCIR \cite{DBLP:conf/cvpr/HouPLWL19}: This framework, initially developed for image classification, has been adapted for CNER \cite{zheng2022distilling}. Like PODNet, it focuses on mitigating catastrophic forgetting. Its loss function comprises three parts: (1) cross-entropy loss for new entity types; (2) distillation loss between features from the previous and current models; and (3) margin-ranking loss for samples containing previously learned entity types.

\item ST \cite{DBLP:conf/wacv/RosenbergHS05,DBLP:journals/corr/abs-1909-08383}: This method uses the pre-existing model to label non-entity tokens with their previous entity types. The updated model is then trained on new data, which includes annotations for all previously encountered entity types. The objective is to minimize cross-entropy loss on this data to enhance model performance.

\item ExtendNER \cite{monaikul2021continual}: ExtendNER applies KD to CNER. It calculates cross-entropy loss for entity type tokens and KL divergence loss for non-entity tokens. During training, ExtendNER minimizes the combined cross-entropy and KL divergence losses.

\item CFNER \cite{zheng2022distilling}: Building upon ExtendNER, CFNER introduces a causal framework for CNER, focusing on distilling causal effects within non-entity type tokens. Initially, the old model identifies non-entity tokens related to previous entity types for KD, while curriculum learning helps mitigate recognition errors.

\item {IS3 \cite{qiu2024incremental}: IS3 addresses two key semantic shifts: mislabeling old entities as non-entities and labeling non-entities or old entities as new entities. It applies KD to preserve recognition of old entities and uses debiased training to reduce bias toward new entities.}

\item CPFD \cite{zhang2023continual}: CPFD tackles two major challenges in CNER: catastrophic forgetting and the semantic shift problem associated with the non-entity type. To mitigate catastrophic forgetting, CPFD introduces a pooled feature distillation loss, which balances stability and adaptability from a representation perspective. At the same time, CPFD employs a pseudo-labeling strategy to address the issue of semantic shift.

\end{itemize}

\subsection{Implementation Specifics}
\label{details}

Consistent with previous CNER methods~\cite{monaikul2021continual, xia2022learn, zheng2022distilling, zhang2023continual}, we use the ``BIO" tagging scheme across all three datasets. This scheme labels each entity type with two tags: B-type (for the beginning of an entity) and I-type (for the interior of an entity).
Our NER model employs the bert-base-cased~\cite{kenton2019bert} encoder, which has a depth of $12$ layers and $12$ attention heads. A fully-connected layer is used as the classifier. We implement the model using PyTorch~\cite{paszke2019pytorch} with the BERT Huggingface implementation~\cite{wolf2019huggingface}. For each setting, if PG=$1$, we train the model for $10$ epochs; otherwise, for $20$ epochs. The batch size, learning rate, and balancing weight $\lambda$ are set to $8$, $4\times 10^{-4}$, and $2$, respectively. All experiments are conducted on an NVIDIA V100 GPU with $32$GB of memory, and each experiment is repeated \textbf{five times} to ensure statistical significance.

\section{Experimental Results}
\label{ex_res}

To illustrate the superiority and efficacy of our SPT method, we conducted extensive experiments to address the following research questions (RQ):

\begin{itemize}
\item \textbf{RQ1}: How does SPT's quantitative performance compare to that of competitive baselines?
\item \textbf{RQ2}: What is the impact of SPT's key components?
\item \textbf{RQ3}: How do key hyper-parameters affect SPT performance?
\item \textbf{RQ4}: How does SPT's qualitative performance measure up against competitive baselines?

\item \textbf{RQ5}: How does SPT maintain stability with varying learning orders of entity types and when employing larger language models as its encoder backbone?

\item \textbf{RQ6}: {How is the runtime efficiency of SPT?}

\item \textbf{RQ7}: {How does SPT perform on Chinese or domain-specific NER datasets?}

\end{itemize}

\subsection{Main Experiments (RQ1)}

In this subsection, we carried out extensive experiments using the CoNLL2003 \cite{sang1837introduction}, I2B2 \cite{murphy2010serving}, and OntoNotes5 \cite{hovy2006ontonotes} datasets under ten different CNER settings to evaluate the quantitative performance of our SPT method against seven competitive baselines. The results are shown in Tables \ref{tab:main_result_1_full}, \ref{tab:main_result_2_full}, and \ref{tab:main_result_3_full}, which display the average Ma-F1 scores for old, new, and all entity types across all CNER steps for each setting, along with the average Mi-F1 score for all entity types across all steps. To ensure statistical reliability, each experiment was repeated five times.

\begin{table*}[t]
\centering
\caption{Comparisons with baselines on the OntoNotes5~\cite{hovy2006ontonotes} dataset. The \textcolor{deepred}{\textbf{red}} indicates the highest result, while the \textcolor{blue}{\textbf{blue}} signifies the second highest result. $*$ indicates results obtained from our re-implementation based on publicly available code. All other baseline results are directly cited from CPFD~\cite{zhang2023continual} or their original papers.}

\resizebox{0.98\linewidth}{!}{%
\begin{tabular}{@{}ccccccccc@{}}
\toprule
&\multicolumn{4}{c}{FG-1-PG-1} & \multicolumn{4}{c}{FG-2-PG-2}  \\ \cmidrule(l){2-5} \cmidrule(l){6-9} 
&\multicolumn{1}{c}{Old Entity Types} & \multicolumn{1}{c}{New Entity Types}&\multicolumn{2}{c}{All Entity Types}& \multicolumn{1}{c}{Old Entity Types} & \multicolumn{1}{c}{New Entity Types}&\multicolumn{2}{c}{All Entity Types} \\
\cmidrule(l){4-5} \cmidrule(l){8-9}
 \multirow{-4}{*}{Baselines} & Avg. Ma-F1   & Avg. Ma-F1  & Avg. Mi-F1  & Avg. Ma-F1  & Avg. Ma-F1   & Avg. Ma-F1  &Avg. Mi-F1  &Avg. Ma-F1  \\\midrule
 
 FT &-- &-- & 15.27±0.26 & 10.86±1.11 &-- & --&25.85±0.11 & 20.55±0.24 \\

{$\text{FT}^{*}$} &0.49±0.03 &53.49±0.46 & 15.06±0.05 & 10.64±0.08 &1.76±0.16 & 61.32±0.17 &25.80±0.02 & 20.35±0.14 \\
 
 PODNet~\cite{DBLP:conf/eccv/DouillardCORV20} &-- &-- & 9.06±0.56 & 8.36±0.57 &-- &-- & 34.67±1.08 & 24.62±0.85 \\

{$\text{PODNet}^{*}$}~\cite{DBLP:conf/eccv/DouillardCORV20} &6.40±0.33 &6.32±0.17 & 9.45±0.25 & 9.40±0.29 &10.88±0.45 &14.36±0.27 & 20.16±0.58 & 17.47±0.26 \\

 LUCIR~\cite{DBLP:conf/cvpr/HouPLWL19} &-- &-- & 28.18±1.15 & 21.11±0.84 &-- &-- & 64.32±1.79 & 43.53±1.11 \\

{$\text{LUCIR}^{*}$}~\cite{DBLP:conf/cvpr/HouPLWL19} &26.22±11.66 &37.73±15.97 & 45.97±18.25 & 30.21±11.52 &53.03±0.44 &58.33±0.35 & 72.87±0.45 & 54.49±0.34 \\ 

ST~\cite{DBLP:conf/wacv/RosenbergHS05} &-- &-- & 50.71±0.79 & 33.24±1.06 &-- &-- & 68.93±1.67 & 50.63±1.66\\

{$\text{ST}^{*}$}~\cite{DBLP:conf/wacv/RosenbergHS05} &28.10±1.25 &40.41±0.84 & 50.31±0.74 & 32.11±1.17 &49.52±1.10 & 54.03±0.71 & 70.01±0.47 & 51.47±0.91\\

 ExtendNER~\cite{monaikul2021continual} &-- &-- & 50.53±0.86 & 32.84±0.84 &-- &-- & 67.61±1.53 & 49.26±1.49 \\

  $\text{ExtendNER}^{*}$~\cite{monaikul2021continual} & 29.63±1.14 &41.11±1.07 & 51.36±0.77 & 33.38±0.98 & 44.90±6.49 & 51.97±2.87 & 63.03±9.39 & 47.64±5.15 \\

 CFNER~\cite{zheng2022distilling} &-- &-- & 58.94±0.57 & 42.22±1.10 &-- &-- & 72.59±0.48 & 55.96±0.69 \\

 $\text{CFNER}^{*}$~\cite{zheng2022distilling} & 37.74±1.77 & 55.01±0.20 & 58.44±0.71 & 41.75±1.51 & 52.70±0.26 & 60.54±0.99 & 72.10±0.31 & 55.02±0.35  \\

{IS3}~\cite{qiu2024incremental} &-- &-- & -- & \textcolor{blue}{\textbf{54.65±0.84}} &-- & --& -- & \textcolor{blue}{\textbf{58.25±0.56}}  \\

 CPFD~\cite{zhang2023continual} &-- &-- & 66.73±0.70 & 54.12±0.30 &-- &-- & \textcolor{blue}{\textbf{74.33±0.30}} & 57.75±0.35 \\

 $\text{CPFD}^{*}$~\cite{zhang2023continual} & \textcolor{blue}{\textbf{53.31±0.43}} & \textcolor{blue}{\textbf{62.33±0.61}} & \textcolor{blue}{\textbf{66.88±0.27}} & 54.20±0.36 & \textcolor{blue}{\textbf{55.16±0.57}} & \textcolor{blue}{\textbf{64.43±0.82}} & 74.03±0.17 & 57.45±0.43  \\

 \midrule

 \textbf{SPT (Ours)} &\textcolor{deepred}{\textbf{57.85±0.48}} & \textcolor{deepred}{\textbf{64.30±0.86}} & \textcolor{deepred}{\textbf{69.09±0.82}} & \textcolor{deepred}{\textbf{58.02±0.29}} & \textcolor{deepred}{\textbf{61.55±0.71}} & \textcolor{deepred}{\textbf{64.78±0.52}} & \textcolor{deepred}{\textbf{76.55±0.41}} & \textcolor{deepred}{\textbf{62.14±0.63}} \\
 
\textbf{Imp.} & $\Uparrow$\textbf{4.54}& $\Uparrow$\textbf{1.97}& $\Uparrow$\textbf{2.21}& $\Uparrow$\textbf{3.37}& $\Uparrow$\textbf{6.39}& $\Uparrow$\textbf{0.35}& $\Uparrow$\textbf{2.22}& $\Uparrow$\textbf{3.89} \\

\midrule
&\multicolumn{4}{c}{FG-8-PG-1} & \multicolumn{4}{c}{FG-8-PG-2}  \\ \cmidrule(l){2-5} \cmidrule(l){6-9} 
&\multicolumn{1}{c}{Old Entity Types} & \multicolumn{1}{c}{New Entity Types}&\multicolumn{2}{c}{All Entity Types}& \multicolumn{1}{c}{Old Entity Types} & \multicolumn{1}{c}{New Entity Types}&\multicolumn{2}{c}{All Entity Types} \\
\cmidrule(l){4-5} \cmidrule(l){8-9}
 \multirow{-4}{*}{Baselines} & Avg. Ma-F1   & Avg. Ma-F1  & Avg. Mi-F1  & Avg. Ma-F1  & Avg. Ma-F1   & Avg. Ma-F1  &Avg. Mi-F1  &Avg. Ma-F1  \\\midrule
 
FT &-- &-- & 17.63±0.57 & 12.23±1.08 &-- &-- & 29.81±0.12 & 20.05±0.16 \\

{$\text{FT}^{*}$} &1.67±0.21 & 58.84±0.31 & 17.72±0.13 & 12.28±0.18 &0.23±0.07 &64.52±0.19 & 30.13±0.07 & 20.53±0.06 \\
 
 PODNet~\cite{DBLP:conf/eccv/DouillardCORV20} &-- &-- & 29.00±0.86 & 20.54±0.91 &-- &-- & 37.38±0.26 & 25.85±0.29 \\

{$\text{PODNet}^{*}$}~\cite{DBLP:conf/eccv/DouillardCORV20} &18.15±0.23 & 16.08±0.38 & 28.66±0.20 & 22.77±0.22 & 23.36±0.43 & 19.96±0.19 & 41.28±0.61 & 31.01±0.31 \\

 LUCIR~\cite{DBLP:conf/cvpr/HouPLWL19} &-- &-- & 66.46±0.46 & 46.29±0.38 &-- &-- & 76.17±0.09 & 55.58±0.55 \\

{$\text{LUCIR}^{*}$}~\cite{DBLP:conf/cvpr/HouPLWL19} &50.50±0.54 &48.48±0.50 & 75.01±0.59 & 52.40±0.48 &56.54±0.40 & 60.95±0.33 & 80.89±0.11 & 59.90±0.30 \\

 ST~\cite{DBLP:conf/wacv/RosenbergHS05} &-- &-- & 73.59±0.66 & 49.41±0.77 &-- &-- & 77.07±0.62 & 53.32±0.63\\

{$\text{ST}^{*}$}~\cite{DBLP:conf/wacv/RosenbergHS05} & 48.57±0.72& 46.26±0.65 & 74.12±0.70 & 50.68±0.65 &50.36±0.75 & 56.82±0.72 & 78.53±0.43 & 55.06±0.60\\

  ExtendNER~\cite{monaikul2021continual} &-- &-- & 73.12±0.93 & 49.55±0.90 &-- &-- & 76.85±0.77 & 54.37±0.57 \\

  $\text{ExtendNER}^{*}$~\cite{monaikul2021continual} & 48.73±0.63 & 46.09±0.56 & 73.65±0.19 & 50.55±0.56 & 51.11±0.70 & 57.41±0.67 & 77.86±0.10 & 55.21±0.51 \\

 CFNER~\cite{zheng2022distilling} & -- & -- & 78.92±0.58 & 57.51±1.32 &-- &-- & 80.68±0.25 & 60.52±0.84  \\

 $\text{CFNER}^{*}$~\cite{zheng2022distilling} & 57.28±0.49 & 59.73±0.60 & 78.25±0.33 & 58.64±0.42 & 58.03±0.47 & 65.01±0.72 & 80.09±0.37 & 61.06±0.37  \\

{IS3}~\cite{qiu2024incremental} &-- &-- & -- & 61.44±0.11 &-- & --& -- & 66.01±0.74  \\

 CPFD~\cite{zhang2023continual} & -- & -- & \textcolor{blue}{\textbf{81.87±0.47}} & 65.52±1.05 &-- &-- & 83.38±0.18 & 66.27±0.75  \\

 $\text{CPFD}^{*}$~\cite{zhang2023continual} & \textcolor{blue}{\textbf{64.65±0.71}} & \textcolor{blue}{\textbf{66.66±0.32}} & 81.76±0.14 & \textcolor{blue}{\textbf{65.59±0.61}} & \textcolor{blue}{\textbf{64.08±0.45}} & \textcolor{blue}{\textbf{69.85±0.43}} & \textcolor{blue}{\textbf{83.43±0.12}} & \textcolor{blue}{\textbf{66.42±0.34}}  \\

 \midrule

 \textbf{SPT (Ours)} & \textcolor{deepred}{\textbf{66.82±0.37}} & \textcolor{deepred}{\textbf{69.10±0.75}} & \textcolor{deepred}{\textbf{84.42±0.43}} & \textcolor{deepred}{\textbf{67.39±0.34}} & \textcolor{deepred}{\textbf{66.16±0.64}} & \textcolor{deepred}{\textbf{71.87±0.89}} & \textcolor{deepred}{\textbf{86.53±0.12}} & \textcolor{deepred}{\textbf{67.78±0.53}} \\
 
\textbf{Imp.} & $\Uparrow$\textbf{2.17}& $\Uparrow$\textbf{2.44}& $\Uparrow$\textbf{2.55}& $\Uparrow$\textbf{1.80}& $\Uparrow$\textbf{2.08}& $\Uparrow$\textbf{2.02}& $\Uparrow$\textbf{3.10}& $\Uparrow$\textbf{1.36} \\

\bottomrule
\end{tabular}
}
\label{tab:main_result_3_full}
\end{table*}

\begin{figure*}[tbp]
\centering
  \includegraphics[width=1.0\linewidth]{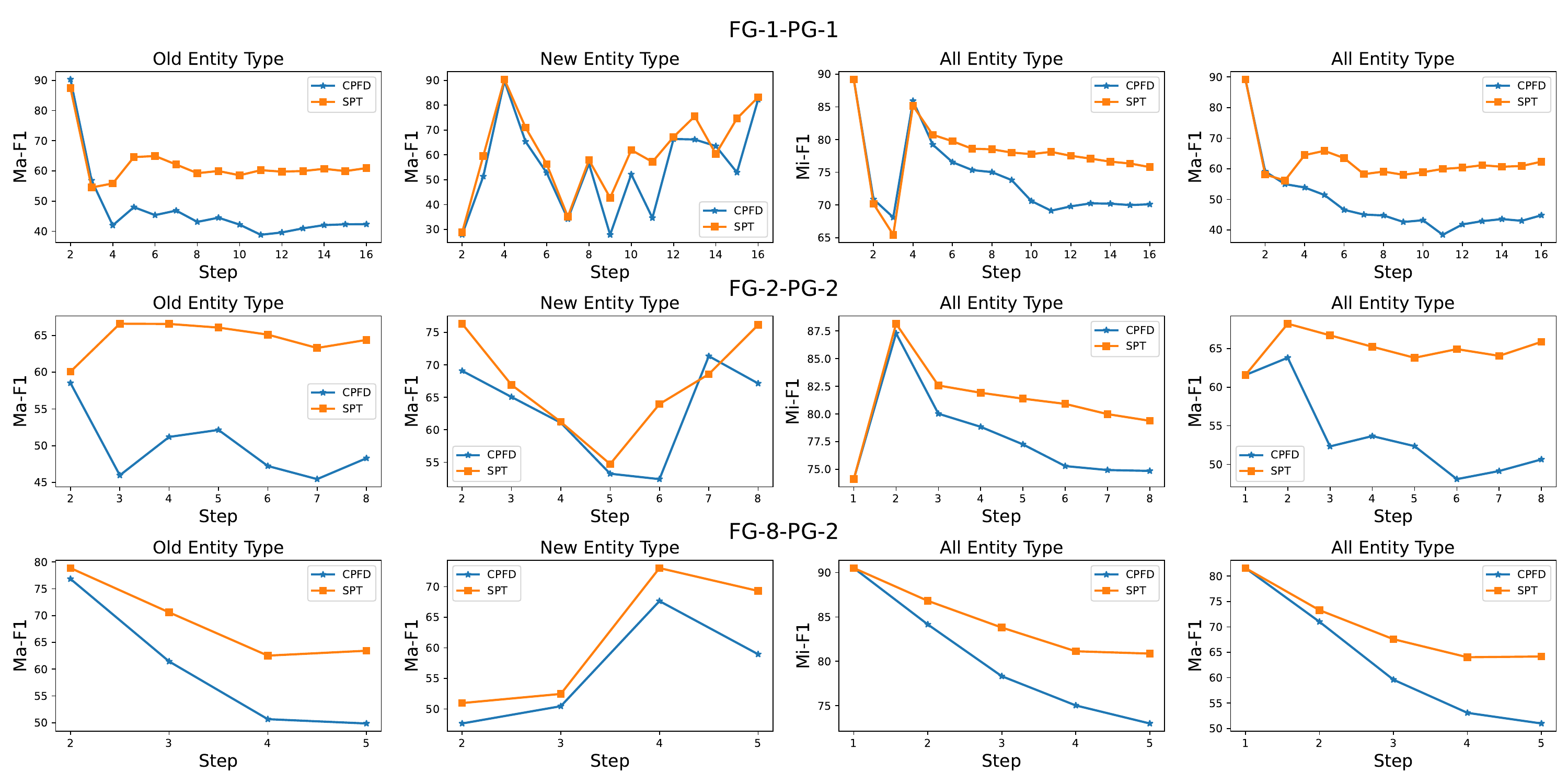}
    \caption{{Comparison of the step-wise Ma-F1 scores for old, new, and all entity types, along with the step-wise Mi-F1 scores for all entity types, using the FG-1-PG-1, FG-2-PG-2, and FG-8-PG-2 settings of the I2B2 dataset \cite{murphy2010serving}. Our SPT method consistently outperforms the previous SOTA CNER baseline, CPFD \cite{zhang2023continual}, in most step-wise evaluations.}}
    \label{fig:step}
\end{figure*}

As illustrated in Table \ref{tab:main_result_1_full}, for the FG-1-PG-1 and FG-2-PG-1 settings of the CoNLL dataset \cite{sang1837introduction}, our SPT method demonstrated average improvements in Mi-F1 scores of $1.98$ and $1.46$, and in Ma-F1 scores of $1.51$ and $1.50$, respectively, for all entity types at each step compared to the previous SOTA CPFD method. For old entity types at each step, the SPT method showed average improvements of $2.22$ and $1.59$ in Ma-F1 scores, respectively. Additionally, for new entity types at each step, our method achieved average improvements of $2.15$ and $1.27$ in Ma-F1 scores, respectively.

As illustrated in Table \ref{tab:main_result_2_full}, for the FG-1-PG-1, FG-2-PG-2, FG-8-PG-1, and FG-8-PG-2 settings of the I2B2 dataset \cite{murphy2010serving}, our SPT method demonstrated average improvements in Mi-F1 scores of $3.62$, $2.87$, $1.06$, and $3.58$, and in Ma-F1 scores of $13.23$, $9.00$, $4.25$, and $5.09$, respectively, for all entity types at each step compared to the previous SOTA CPFD method. For old entity types at each step, the SPT method showed average improvements of $14.90$, $14.75$, $7.57$, and $9.16$ in Ma-F1 scores, respectively. Additionally, for new entity types at each step, our method achieved average improvements of $6.54$, $4.08$, $3.94$, and $5.27$ in Ma-F1 scores, respectively. {Finally, our SPT method also outperforms the latest IS3 \cite{qiu2024incremental} approach in these four settings, with improvements of $5.45$, $3.21$, $2.06$, and $7.10$ respectively in Avg. Ma-F1 scores across all entity types.}

As illustrated in Table \ref{tab:main_result_3_full}, for the FG-1-PG-1, FG-2-PG-2, FG-8-PG-1, and FG-8-PG-2 settings of the OntoNotes5 dataset \cite{hovy2006ontonotes}, our SPT method demonstrated average improvements in Mi-F1 scores of $2.21$, $2.22$, $2.55$, and $3.10$, and in Ma-F1 scores of $3.82$, $4.39$, $1.80$, and $1.36$, respectively, for all entity types at each step compared to the previous SOTA CPFD method. For old entity types at each step, the SPT method showed average improvements of $4.54$, $6.39$, $2.17$, and $2.08$ in Ma-F1 scores, respectively. Additionally, for new entity types at each step, our method achieved average improvements of $1.97$, $0.35$, $2.44$, and $2.02$ in Ma-F1 scores, respectively. 
{Finally, our SPT method also outperforms the latest IS3 \cite{qiu2024incremental} approach in these four settings, with improvements of $3.37$, $3.89$, $5.95$, and $1.77$ respectively in Avg. Ma-F1 scores across all entity types.}

These findings quantitatively validate the superiority of our SPT method in comparison to baselines, showcasing its capacity to develop a robust CNER model. The results suggest enhanced resistance to catastrophic forgetting by achieving a suitable balance between stability (\emph{i.e.}, preserving prior knowledge) and plasticity (\emph{i.e.}, acquiring new knowledge).

\subsection{Step-wise Comparisons (RQ1)}

{As shown in Figure \ref{fig:step}, we present step-wise comparison results to comprehensively evaluate the effectiveness of our SPT method in CNER scenarios. The results demonstrate that SPT consistently outperforms the previous SOTA baseline, CPFD \cite{zhang2023continual}, across most step-wise evaluations. This superior performance is particularly evident under various experimental configurations, including the FG-1-PG-1, FG-2-PG-2, and FG-8-PG-2 settings on the I2B2 dataset \cite{murphy2010serving}. These findings highlight the robustness of SPT in sustaining high performance and adaptability, even as it learns to recognize new and evolving entity types. The notable improvement over CPFD further demonstrates SPT's potential to serve as a new benchmark for CNER tasks, especially in complex, real-world settings.}

\begin{table}[tbp]
  \centering
\caption{The ablation study of our SPT on the I2B2 \cite{murphy2010serving} and OntoNotes5 \cite{hovy2006ontonotes} datasets under the FG-1-PG-1 setting. Compared to our full SPT method, all ablation variants show a significant decline in CNER performance. These results underscore the critical role each component plays in collaboratively addressing CNER. VWF and WSM denote Vanilla Weight Fusion and Weight-guide Selective Mechanism, respectively. \textbf{Bold} denotes the best results.}
  \resizebox{1.0\linewidth}{!}{
    \begin{tabular}{lcccc}
    \toprule
    \multirow{2}[4]{*}{Methods} & \multicolumn{2}{c}{I2B2 \cite{murphy2010serving}} & \multicolumn{2}{c}{OntoNotes5 \cite{hovy2006ontonotes}}  \\
\cmidrule{2-5}          & Avg. Mi-F1 & Avg. Ma-F1 & Avg. Mi-F1 & Avg. Ma-F1  \\
    \midrule
    
    \textbf{SPT (Ours)}  & \textbf{75.95±0.61}  & \textbf{53.07±1.26}  & \textbf{69.09±0.82}  & \textbf{58.02±0.29}  \\

    \midrule

    \ \ \ w/ $\mathcal{L}_\text{KD}$ (Equation (\ref{kd})) &74.14±1.08& 49.96±1.16& 66.28±1.15& 55.55±0.87\\
    \ \ \ w/ $\mathcal{L}_\text{PKD-lax}$ (Equation (\ref{lax}))&71.86±1.02& 48.75±1.29& 64.59±0.61& 53.76±0.82\\
    \ \ \ w/o $\mathcal{L}_\text{PKD}$ (Equation (\ref{pkd}))&70.70±0.97& 47.38±0.96& 62.98±0.83& 51.67±1.10\\
  \midrule

    \ \ \ w/ VWF (Equation (\ref{wf_eq})) &74.86±0.94& 50.27±1.17& 67.83±0.97& 56.23±0.72\\ 
    \ \ \ w/o WSM (Equation (\ref{fused})) & 74.19±0.95  & 48.34±1.45  & 66.73±0.70  & 54.12±0.30\\

      \midrule
    \ \ \ {w/o Pseudo-labeling Threshold} &72.91±0.76 & 50.95±1.24& 66.67±0.61& 55.40±0.93 \\
    
    \bottomrule
    \end{tabular}%
    }

  \label{tab:ablation_study}%
\end{table}%

\subsection{Ablation Study (RQ2)}

This subsection evaluates the effectiveness of the individual components in our SPT method through ablation studies, as shown in Table~\ref{tab:ablation_study}. We observe that replacing our $\mathcal{L}_\text{PKD}$ loss (Equation (\ref{pkd})) with either a stricter $\mathcal{L}_\text{KD}$ loss (Equation (\ref{kd})) or a more lenient $\mathcal{L}_\text{PKD-lax}$ loss (Equation (\ref{lax})) generally leads to reduced CNER performance. This reduction occurs because these variations either exclude pooling or implement more aggressive pooling, resulting in either excessive stability or plasticity in the CNER model. Our $\mathcal{L}_\text{PKD}$ loss achieves a balanced approach between stability and plasticity, which helps mitigate catastrophic forgetting; without this balance, poor CNER performance will result.
Additionally, substituting our Weight-guided Selective Mechanism (WSM) (Equation (\ref{fused})) with a Vanilla Weight Fusion strategy (VWF) (Equation (\ref{wf_eq}))—which uniformly constrains all weights of the old model to counteract catastrophic forgetting—results in suboptimal performance. This highlights the importance of selectively integrating crucial weights from the previous model. Our WSM provides a better balance between old and new knowledge from a weight perspective, and without this mechanism, poor CNER performance will result.

{Moreover, SPT w/o Pseudo-labeling Threshold refers to directly using the raw predictions from the old model without applying the entropy-based confidence threshold to filter out errors in the naive pseudo-labeling process. The results show that incorporating the entropy-based threshold effectively reduces noise and yields higher-quality pseudo labels.}

\begin{figure*}[tbp]
\centering
  \includegraphics[width=1.0\textwidth]{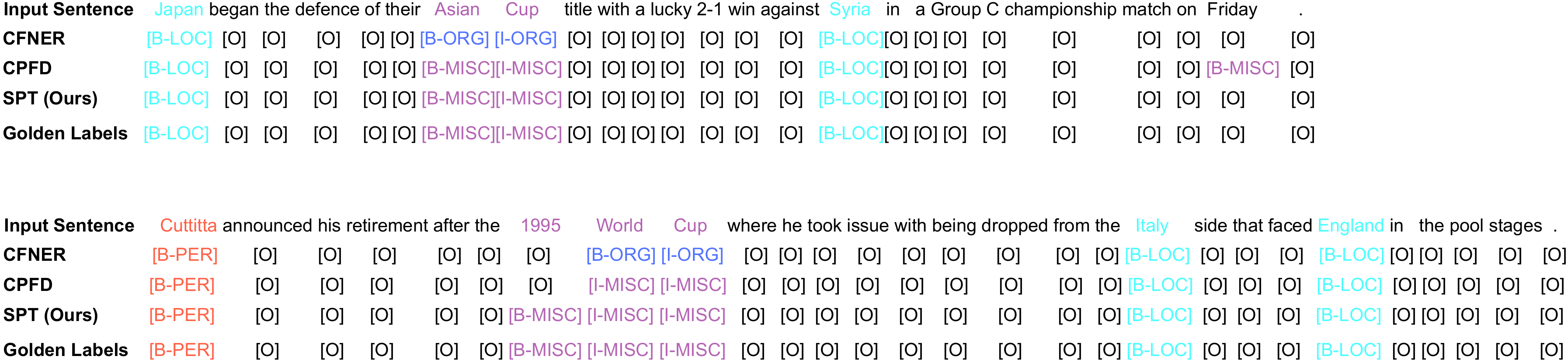}
\caption{Two predicted NER cases sampled from the CoNLL2003 \cite{sang1837introduction} test set. \textbf{B-} and \textbf{I-} denote the beginning and inside of named entities, respectively. [O], \textcolor{cyan}{[LOC]}, \textcolor{purple}{[MISC]}, \textcolor{blue}{[ORG]}, and \textcolor{red}{[PER]} represent non-entity, \textcolor{cyan}{\textbf{Location}}, \textcolor{purple}{\textbf{Other entity types}}, \textcolor{blue}{\textbf{Organization}}, and \textcolor{red}{\textbf{Person}}, respectively. All prediction results are from the final step of the FG-1-PG-1 setting. These visualized NER cases qualitatively demonstrate the superiority and effectiveness of our SPT method.}
\label{fig:case}
\end{figure*}

\begin{table}[tbp]
  \centering
\caption{The hyper-parameter analysis of $\lambda$ on the I2B2 \cite{murphy2010serving} dataset under the FG-8-PG-2 setting. \textbf{Bold} represents the best results.}
  \resizebox{0.85\linewidth}{!}{
    \begin{tabular}{ccc}
    \toprule
    $\lambda$ in Equation (\ref{eq:complete_loss}) & Avg. Mi-F1 & Avg. Ma-F1  \\
    \midrule

0.5 & 83.69±0.70 &69.61±0.88 \\

1.0 & 83.54±1.06&69.39±0.97 \\

2.0 & \textbf{84.63±0.31} & \textbf{70.13±0.73} \\

5.0 & 80.38±1.11&66.46±1.01 \\
   
    \bottomrule
    \end{tabular}%
    } 
  \label{hyper}%
\end{table}%

\subsection{Hyper-parameter Analysis (RQ3)}

We conduct a thorough analysis of the hyper-parameter $\lambda$ on the I2B2 dataset within the FG-8-PG-2 setting. This hyper-parameter, as specified in Equation (\ref{eq:complete_loss}), plays a crucial role in balancing the loss, with detailed results presented in Table \ref{hyper}. Our findings indicate that the model achieves optimal performance when $\lambda$ is set to $2$. However, it is important to note that performance significantly deteriorates when $\lambda$ is increased beyond this value, suggesting that overemphasizing certain aspects of the loss can be detrimental.

It is also worth mentioning that we did not perform an exhaustive search for the best hyper-parameters; instead, we used the default setting of $\lambda = 2$ throughout our experiments. This choice, while sufficient for our purposes, implies that there may be room for further improvement. A more meticulous tuning of hyper-parameters could potentially yield even better performance on specific datasets and experimental settings. This observation highlights the importance of hyper-parameter optimization in fine-tuning model performance and suggests that future research could explore this avenue to achieve superior results.

\subsection{Case Study (RQ4)}

In this subsection, we provide a detailed qualitative evaluation of our SPT method, comparing its performance against leading baseline models such as CFNER \cite{zheng2022distilling} and CPFD \cite{zhang2023continual}. To illustrate this comparison, we include a visual qualitative analysis based on the FG-1-PG-1 setting of the CoNLL2003 dataset \cite{sang1837introduction}, as depicted in Figure \ref{fig:case}. These visual assessments clearly demonstrate the superior ability of our SPT method to learn new entity types in a sequential manner. This highlights its effectiveness and robustness within the continual learning framework, where the model must continuously adapt to new information without forgetting previously learned knowledge.

The comparison not only underscores SPT's advanced capacity for handling new and emerging entities but also showcases its advantages over other SOTA methods. By effectively managing the balance between retaining existing knowledge and integrating new information, SPT proves to be a powerful tool in the ongoing development of more adaptable and resilient NER systems. This qualitative analysis further supports the previous quantitative results, providing a comprehensive view of SPT's strengths in real-world, dynamic learning scenarios.

\begin{figure*}[t]
\centering
  \includegraphics[width=1.0\linewidth]{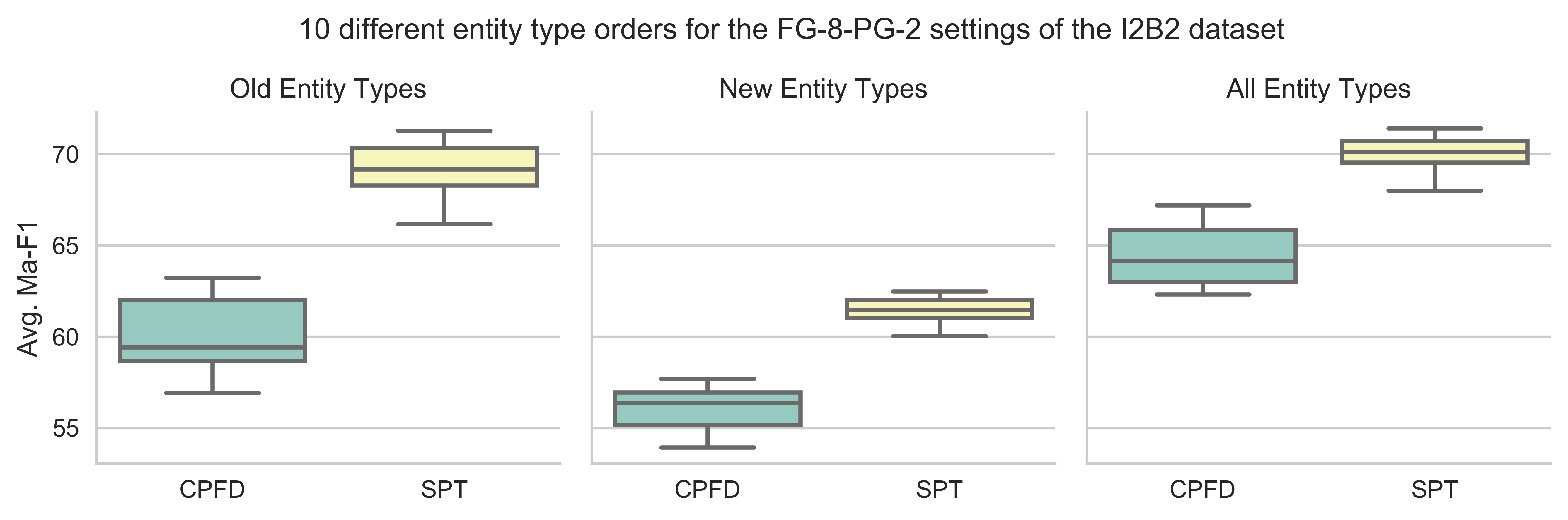} 
   \caption{The boxplots illustrate the average Ma-F1 scores for old, new, and all entity types across CNER steps for $10$ random entity type orders. SPT demonstrates significantly better performance and greater stability compared to CPFD.}
\label{fig:order}
\end{figure*}

\subsection{Stability with Respect to Entity Type Orders (RQ5)}

Recent studies have revealed that existing continual learning methods can be prone to instability, with the sequence in which entity types are introduced having a significant impact on overall performance \cite{DBLP:conf/cvpr/DouillardCDC21,kim2019incremental}. This presents a challenge in real-world scenarios, where the optimal order of entity types is not known beforehand. Consequently, an ideal CNER method should demonstrate consistent performance, irrespective of the order in which entity types are learned.

In previous experiments, the order of entity types has been kept consistent, typically following an alphabetical sequence as outlined in \cite{zheng2022distilling,zhang2023continual}. To further evaluate the robustness of our method, we conducted an experiment where we applied $10$ random permutations of the entity type order under the FG-8-PG-2 setting of the I2B2 dataset \cite{murphy2010serving}. The results, presented in Figure \ref{fig:order}, use boxplots to depict the average Ma-F1 scores for old, new, and all entity types across all CNER steps.

As shown in the figure, the boxplot for our SPT method displays higher average values, a more concentrated data distribution, and a narrower range of variation compared to the CPFD baseline. This indicates that SPT not only outperforms CPFD but also offers greater stability and reliability, making it a more robust solution for continual learning tasks where the order of entity types cannot be predetermined. The consistency of SPT across different permutations underscores its effectiveness in handling the dynamic and unpredictable nature of real-world data environments.

\begin{table}[tbp]
  \centering
\caption{Performance on the I2B2 \cite{murphy2010serving} dataset under the FG-8-PG-2 setting, where the encoder backbone is replaced by larger PLMs. 
The I2B2 dataset contains 16 entity types in total. 
In the FG-8-PG-2 setting, the model learns 8 entity types in the first step (FG), and then incrementally learns 2 new types in each subsequent step (PG), resulting in 5 continual learning steps. 
Evaluation metrics include Avg. Mi-F1 and Avg. Ma-F1, computed by averaging the Mi-F1 and Ma-F1 scores across all learned entity types at each step, and then averaging over all steps. \textbf{Bold} indicates the best performance.}
  \resizebox{0.85\linewidth}{!}{
    \begin{tabular}{ccc}
    \toprule
   Encoder Backbones & Avg. Mi-F1 & Avg. Ma-F1  \\
    \midrule
    bert-base-cased \cite{kenton2019bert} & 84.63±0.31 & 70.13±0.73\\
    bert-large-cased \cite{kenton2019bert} & 85.51±0.93 & 71.80±1.04\\    
    roberta-large \cite{liu2019roberta}  & \textbf{86.15±0.89} & \textbf{72.26±1.12} \\
    \bottomrule
    \end{tabular}
    }
  \label{tab:encoder}
\end{table}

\subsection{Using Larger Encoder Backbones (RQ5)}

In our earlier experiments, we primarily used the bert-based-cased model as the encoder backbone \cite{kenton2019bert}. In this section, we investigate the impact of employing larger PLMs as the encoder, specifically exploring the performance of bert-large-cased \cite{kenton2019bert} and roberta-large \cite{liu2019roberta}. The results obtained from the I2B2 dataset \cite{murphy2010serving} under the FG-8-PG-2 setting are summarized in Table \ref{tab:encoder}. Our findings reveal that substituting the encoder backbone with these larger PLMs consistently yields notable performance improvements.

The enhanced results demonstrate the effectiveness of using more powerful PLMs, which likely benefit from their greater capacity to capture complex patterns and relationships within the data. This suggests that scaling up the model size can lead to more accurate and robust outcomes in CNER tasks, particularly in scenarios involving complex and evolving datasets. The consistent gains observed with larger models highlight the potential for further advancements by exploring even more sophisticated PLMs in future research.

\begin{table}[tbp]
  \centering
\caption{Efficiency comparison between SPT and CPFD \cite{zhang2023continual} under the FG-8-PG-1 setting on the I2B2 \cite{murphy2010serving} dataset. 
The I2B2 dataset includes 16 entity types. 
In the FG-8-PG-1 setting, the model learns 8 entity types in the initial step (FG), followed by 1 new type per step in each of the subsequent steps (PG), resulting in 9 continual learning steps. 
Evaluation metrics include Avg. Mi-F1 and Avg. Ma-F1, calculated by averaging the Mi-f1 and Ma-f1 scores across all learned entity types at each step, and then averaging over all steps. 
Total runtime denotes the cumulative duration from the second to the final step of continual learning, including both training and inference time. Memory usage refers to the peak GPU memory consumption during training. 
\textbf{Bold} indicates the best results.}
  \resizebox{1.0\linewidth}{!}{
    \begin{tabular}{ccccc}
    \toprule
   Methods & Avg. Mi-F1 $\Uparrow$ & Avg. Ma-F1 $\Uparrow$ &  \makecell{Total Runtime\\(min)} $\Downarrow$& \makecell{Memory Usage\\(MB)} $\Downarrow$ \\
    \midrule
 CPFD \cite{zhang2023continual} &  73.80±1.95& 55.06±1.33& \textbf{1195} & \textbf{4826}\\

 \textbf{SPT (Ours)} &\textbf{75.81±0.42} & \textbf{60.44±0.95} & 1236 & 4830  \\
    \bottomrule
    \end{tabular}
    }
  \label{tab:efficiency}
\end{table}

\subsection{{Efficiency Analysis (RQ6)}}

{We conducted an efficiency analysis of our SPT method and the previous SOTA CPFD \cite{zhang2023continual} under the FG-8-PG-1 setting on the I2B2 \cite{murphy2010serving} dataset, evaluating both runtime (in minutes) and GPU memory usage (in MB). The measured runtime represents the total duration from the second to the final step of continual learning, including both training and inference, as both methods share the same base model and therefore begin from the second step. As shown in Table \ref{tab:efficiency}, SPT incurs a slightly longer toal runtime compared to CPFD, while peak GPU memory usage remains nearly identical between the two. Nonetheless, SPT significantly outperforms CPFD in terms of performance, highlighting the practical efficiency of our approach given its comparable resource requirements.}

\begin{table}[tbp]
  \centering
\caption{{Performance on the CMeEE \cite{zhang2022cblue} dataset under the FG-1-PG-1 CNER setting. \textbf{Bold} indicates the best results.}}
  \resizebox{0.75\linewidth}{!}{
    \begin{tabular}{ccc}
    \toprule
   Methods & Avg. Mi-F1 & Avg. Ma-F1  \\
    \midrule
    CPFD \cite{zhang2023continual} & 48.72±0.85 & 38.42±1.23\\
  
    \textbf{SPT (Ours)}  & \textbf{51.28±1.15} & \textbf{43.66±0.97} \\

    \bottomrule
    \end{tabular}
    }
  \label{tab:chinese_ner}
\end{table}

\subsection{{Generalization to Chinese and Domain-specific Data (RQ7)}}

{We conducted additional experiments on a Chinese biomedical NER dataset, CMeEE \cite{zhang2022cblue}, containing $9$ entity types. To evaluate continual learning performance, we adopted a challenging FG-1-PG-1 setting, where the base model is trained on the first entity type, and one new entity type is introduced at each step, resulting in a total of $9$ steps. Given the Chinese nature of the dataset, we used the bert-base-chinese model as the encoder. As shown in Table \ref{tab:chinese_ner}, our SPT method significantly outperforms the previous SOTA CPFD \cite{zhang2023continual}, demonstrating its strong generalization capability on domain-specific Chinese datasets.}

\section{Conclusion}
\label{conclusion}

In this paper, we introduce a SPT method for CNER that skillfully balances the retention of old knowledge and the acquisition of new knowledge from both representation and weight perspectives. On the representation side, we improve upon the original KD technique by incorporating pooling operations, which help to properly consolidate the representation dimensions and effectively alleviate catastrophic forgetting. From the weight perspective, we develop a weight-guided selective mechanism that strategically integrates critical weights from the old model, specifically those related to previous entity types, into the new model. 
We evaluate the SPT method across ten different CNER settings using three distinct datasets, and the results clearly demonstrate that SPT consistently outperforms previous SOTA methods in all evaluated scenarios. 

Looking ahead, our goal is to extend the application of our approach to other sequence labeling tasks, exploring its potential in a wider range of contexts. Moreover, we plan to investigate the scalability and robustness of the SPT method in more complex continual learning environments, further enhancing its applicability in real-world scenarios.

\bibliographystyle{IEEEtran}
\bibliography{reference}

\begin{IEEEbiography}[{\includegraphics[width=1in,height=1.25in,clip,keepaspectratio]{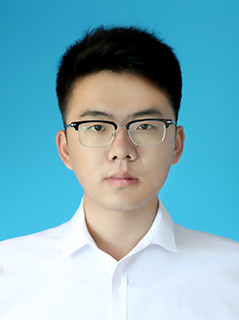}}]{Duzhen Zhang} received his B.Sc. degree from Shandong University in June 2019. He completed his Ph.D. at the Institute of Automation, Chinese Academy of Sciences in June 2024. Since September 2024, he has been a postdoctoral researcher at the Mohamed bin Zayed University of Artificial Intelligence. His current research interests include large language models, continual learning, multi-modal learning, and AI for science.
\end{IEEEbiography}

\begin{IEEEbiography}[{\includegraphics[width=1in,height=1.25in,clip,keepaspectratio]{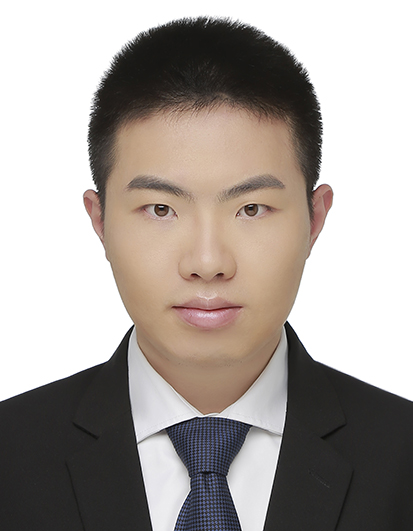}}]{Chenxing Li} received his B.Sc. degree at the North China Electric Power University, China, in 2015. He completed his Ph.D. at the Institute of Automation, Chinese Academy of Sciences in 2020. His current research interests include far-field speech recognition, speech enhancement, and speech separation.
\end{IEEEbiography}

\begin{IEEEbiography}[{\includegraphics[width=1in,height=1.25in,clip,keepaspectratio]{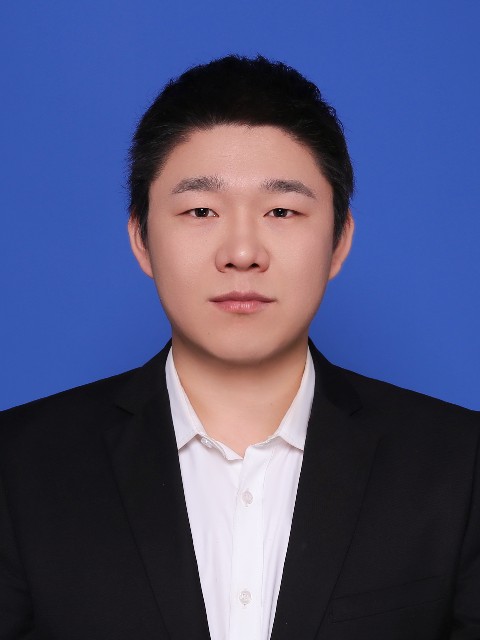}}]{Jiahua Dong} is a postdoctoral researcher in the Mohamed bin Zayed University of Artificial Intelligence, Abu Dhabi, United Arab Emirates. He received the Ph.D. degree from the Shenyang Institute of Automation, Chinese Academy of Sciences in 2024. He visited the Computer Vision Lab, ETH Zurich, Switzerland from April 2022 to August 2022, and Max Planck Institute for Informatics, Germany from September 2022 to January 2023. Before that, he received the B.S. degree from Jilin University in 2017. His current research interests include computer vision, machine learning and medical image analysis.
\end{IEEEbiography}

\begin{IEEEbiography}
[{\includegraphics[width=1in,height=1.25in,clip,keepaspectratio]{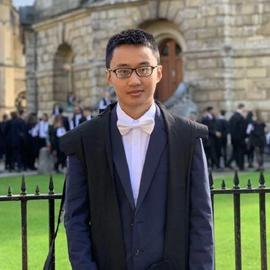}}]{Qi Liu} is an assistant professor at the Department of Computer Science, the University of Hong Kong, and a cofounder of Reka. He earned his Ph.D. in computer science from the University of Oxford. In the past, he obtained a Master of Science degree from the National University of Singapore, and a Bachelor of Engineering degree from Shandong University. His research interests include natural language processing and machine learning. His research is centered on enabling computers to comprehend human language.
\end{IEEEbiography}

\begin{IEEEbiography}[{\includegraphics[width=1in,height=1.25in,clip,keepaspectratio]{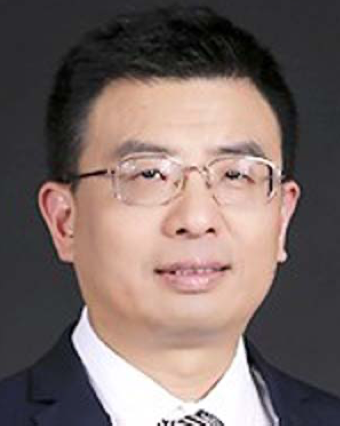}}]{Dong Yu} (Fellow, IEEE) with the Tencent AI Lab as a distinguished scientist and vice general Manager. Prior to joining Tencent in 2017, he was a Principal Researcher with Microsoft Research (Redmond), where he has been since 1998. He has authored or coauthored two monographs and more than 300 papers. His research interests include speech recognition and processing and natural language processing. His works have been widely cited and recognized by the prestigious IEEE Signal Processing Society best transaction paper award in 2013, 2016, 2020, and 2022, the 2021 NAACL best long paper award, 2022 IEEE Signal Processing Magazine best paper award, and 2022 IEEE Signal Processing Magazine best column award. Dr. Dong Yu was the Chair of the IEEE Speech and Language Processing Technical Committee during 2021–2022. He was on the editorial boards of numerous journals and magazines, as well as on the organizing and technical committees of various conferences and workshops. He is currently an ACM/IEEE/ISCA Fellow.
\end{IEEEbiography}

\end{document}